\documentclass[11pt,a4paper]{article}
\usepackage[hyperref]{conll-2019}

\usepackage[utf8]{inputenc}
\usepackage[T1]{fontenc}
\usepackage{times}
\usepackage{mathptmx}	% txfonts
\usepackage[scaled=.8]{beramono}
\usepackage[scaled=.85]{helvet}
\usepackage{MnSymbol}	% must be after mathptmx
\usepackage{latexsym}
\usepackage{microtype}
\usepackage{relsize}
\usepackage{svg}
\usepackage{stfloats}

\usepackage[font=10pt,labelfont=bf,skip=5pt]{caption}

\usepackage[shortlabels]{enumitem} % customizable lists
\setitemize{noitemsep,topsep=0em} %leftmargin=1em,iciteposstemindent=-1em
\setenumerate{noitemsep,leftmargin=0em,itemindent=13pt,topsep=0em}

\usepackage{url}
\usepackage{xcolor}
\usepackage{tikz}
\usepackage{tikz-qtree} % has to go before gb4e
\usepackage{forest}

\useforestlibrary{linguistics}
\forestapplylibrarydefaults{linguistics}

\usepackage{multirow}

\usepackage{arydshln}

\aclfinalcopy % Uncomment this line for the final submission
\usepackage{nameref}
\usepackage{cleveref}

\newcommand{\finalversion}[1]{}
\newcommand{\longversion}[1]{}
\newcommand{\shortlong}[2]{#1}

\crefformat{part}{\S#2#1#3}
\crefformat{chapter}{\S#2#1#3}
\crefformat{section}{\S#2#1#3}
\crefformat{subsection}{\S#2#1#3}
\crefformat{subsubsection}{\S#2#1#3}
\crefformat{paragraph}{\P#2#1#3}
\crefformat{subparagraph}{\P#2#1#3}
\crefmultiformat{section}{\S#2#1#3}{ and~\S#2#1#3}{, \S#2#1#3}{, and~\S#2#1#3}
\crefmultiformat{subsection}{\S#2#1#3}{ and~\S#2#1#3}{, \S#2#1#3}{, and~\S#2#1#3}
\crefmultiformat{subsubsection}{\S#2#1#3}{ and~\S#2#1#3}{, \S#2#1#3}{, and~\S#2#1#3}
\crefmultiformat{paragraph}{\P\P#2#1#3}{ and~#2#1#3}{, #2#1#3}{, and~#2#1#3}
\crefmultiformat{subparagraph}{\P\P#2#1#3}{ and~#2#1#3}{, #2#1#3}{, and~#2#1#3}
\crefrangeformat{section}{\mbox{\S\S#3#1#4--#5#2#6}}
\crefrangeformat{subsection}{\mbox{\S\S#3#1#4--#5#2#6}}
\crefrangeformat{subsubsection}{\mbox{\S\S#3#1#4--#5#2#6}}
\crefrangeformat{paragraph}{\mbox{\P\P#3#1#4--#5#2#6}}
\crefrangeformat{subparagraph}{\mbox{\P\P#3#1#4--#5#2#6}}
\crefname{part}{Part}{Parts}
\Crefname{part}{Part}{Parts}
\crefname{chapter}{ch.}{ch.}
\Crefname{chapter}{Ch.}{Ch.}
\crefname{figure}{figure}{figures}
\crefname{subfigure}{figure}{figures}
\Crefname{subfigure}{Figure}{Figures}
\crefname{appsec}{appendix}{appendices}
\Crefname{appsec}{Appendix}{Appendices}
\crefname{algocf}{algorithm}{algorithms}
\Crefname{algocf}{Algorithm}{Algorithms}
\crefname{enums,enumsi}{example}{examples}
\Crefname{enums,enumsi}{Example}{Examples}
\crefname{}{example}{examples} % lingmacros \toplabel has no internal name for the kind of label
\Crefname{}{Example}{Examples}
\crefformat{enums}{(#2#1#3)}
\crefformat{enumsi}{(#2#1#3)}
\crefrangeformat{enums}{\mbox{(#3#1#4--#5#2#6)}}
\crefrangeformat{enumsi}{\mbox{(#3#1#4--#5#2#6)}}
\crefformat{}{(#2#1#3)}
\crefname{xnumi}{example}{examples} % gb4e
\crefname{xnumi}{example}{examples} % gb4e
\Crefname{xnumii}{Example}{Examples} % gb4e
\Crefname{xnumii}{Example}{Examples} % gb4e
\crefformat{xnumi}{(#2#1#3)} % gb4e
\crefformat{xnumii}{(#2#1#3)} % gb4e
\crefrangeformat{enums}{\mbox{(#3#1#4--#5#2#6)}}
\crefrangeformat{enumsi}{\mbox{(#3#1#4--#5#2#6)}}
\crefrangeformat{xnumi}{\mbox{(#3#1#4--#5#2#6)}} % gb4e
\crefrangeformat{xnumii}{\mbox{(#3#1#4--#5#2#6)}} % gb4e
\crefmultiformat{enumsi}{(#2#1#3}{, #2#1#3)}{, #2#1#3}{, #2#1#3)}
\crefmultiformat{xnumi}{(#2#1#3}{, #2#1#3)}{, #2#1#3}{, #2#1#3)} % gb4e
\crefmultiformat{xnumii}{(#2#1#3}{, #2#1#3)}{, #2#1#3}{, #2#1#3)} % gb4e
\crefrangemultiformat{enumsi}{(#3#1#4--#5#2#6}{, #3#1#4--#5#2#6)}{, #3#1#4--#5#2#6}{, #3#1#4--#5#2#6)}
\crefrangemultiformat{xnumi}{(#3#1#4--#5#2#6}{, #3#1#4--#5#2#6)}{, #3#1#4--#5#2#6}{, #3#1#4--#5#2#6)} % gb4e
\crefrangemultiformat{xnumii}{(#3#1#4--#5#2#6}{, #3#1#4--#5#2#6)}{, #3#1#4--#5#2#6}{, #3#1#4--#5#2#6)} % gb4e

\ifx\creflastconjunction\undefined%
\newcommand{\creflastconjunction}{, and\nobreakspace} % Oxford comma for lists
\else%
\renewcommand{\creflastconjunction}{, and\nobreakspace} % Oxford comma for lists
\fi%

\newcommand*{\Fullref}[1]{\hyperref[{#1}]{\Cref*{#1}: \nameref*{#1}}}
\newcommand*{\fullref}[1]{\hyperref[{#1}]{\cref*{#1}: \nameref{#1}}}
\makeatletter
\renewcommand{\paragraph}{%
  \@startsection{paragraph}{4}%
  {\z@}{.2ex \@plus 1ex \@minus .2ex}{-1em}%
  {\normalfont\normalsize\bfseries}%
}
\makeatother

\usepackage{color}
\usepackage{bm}
\definecolor{orange}{rgb}{1,0.5,0}
\definecolor{mdgreen}{rgb}{0.05,0.6,0.05}
\definecolor{mdblue}{rgb}{0,0,0.7}
\definecolor{dkblue}{rgb}{0,0,0.5}
\definecolor{dkgray}{rgb}{0.3,0.3,0.3}
\definecolor{slate}{rgb}{0.25,0.25,0.4}
\definecolor{gray}{rgb}{0.5,0.5,0.5}
\definecolor{ltgray}{rgb}{0.7,0.7,0.7}

\definecolor{lavender}{rgb}{0.65,0.55,1.0}

\newcommand{\ensuretext}[1]{#1}
\newcommand{\nssmarker}{\ensuretext{\textcolor{magenta}{\ensuremath{^{\textsc{NS}}_{\textsc{S}}}}}}
\newcommand{\jpmarker}{\ensuretext{\textcolor{purple}{\ensuremath{^{\textsc{J}}_{\textsc{P}}}}}}

\newcommand{\arkcomment}[3]{\ensuretext{\textcolor{#3}{[#1 #2]}}}
\newcommand{\nss}[1]{\arkcomment{\nssmarker}{#1}{magenta}}
\newcommand{\jp}[1]{\arkcomment{\jpmarker}{#1}{purple}}
\newcommand{\p}[1]{\textbf{\textsf{#1}}} % preposition type
\newcommand{\lbl}[1]{\textsc{#1}} % class label
\newcommand{\sst}[1]{\lbl{#1}} % supersense tag label
\newcommand{\psst}[1]{\textcolor{mdgreen}{\sst{#1}}} % preposition supersense tag label
\usepackage{linguex}
\crefname{ExNo}{}{}
\Crefname{ExNo}{}{}
\crefname{SubExNo}{}{}
\Crefname{SubExNo}{}{}
\title{Made for Each Other: Broad-coverage Semantic Structures Meet Preposition Supersenses}
\author{Jakob Prange \quad
        Nathan Schneider \\
        Georgetown University \\
        {\tt jakob@cs.georgetown.edu} \\
        {\tt nathan.schneider@georgetown.edu} \\\And
        Omri Abend \\
        The Hebrew University of Jerusalem \\
        {\tt oabend@cs.huji.ac.il}}
\begin{document}
\maketitle

%%%%%%%%%%%%%%%%%%%%%%%%%%%%%%%%%%%%%%%%%%%%%%%%%%%%%%%%%%%%%%%%%%%%%%%%%
\begin{abstract}
%While broad-coverage semantic parsing is demonstrating increasing utility to a variety
%of applications, the range of distinctions it covers is still restricted, and many
%distinctions are still covered by specialized schemes, and not supported by the general ones. 
Universal Conceptual Cognitive Annotation \citep[UCCA;][]{abend2013universal} is a typologically-informed, broad-coverage semantic annotation scheme that describes coarse-grained predicate-argument structure
%and inter-scene linkage phenomena, 
but currently lacks semantic roles. 
We argue that lexicon-free annotation of the semantic roles marked by prepositions, as formulated by \citet{schneider2018comprehensive}, is complementary and suitable for integration within UCCA.
%the UCCA scheme foundational layer with the SNACS
%scheme that addresses the semantic classes of prepositions and case markers.
We show empirically for English that the schemes, though annotated independently, 
are compatible and can be combined in a single semantic graph.
% We advocate a modular design, where specialized annotation schemes are
% integrated as distinct layers on top of UCCA, while obeying unifying design
% principles.
% We show that in the case of the annotation schemes in question,
% integrating the two into a single scheme is beneficial for improving parsing results
% using multi-task learning.\oa{pending on results.}
%Parsing experiments for English test algorithms for predicting this structure jointly or in a pipeline.\nss{or something interesting about experiments}
A comparison of several approaches to parsing the integrated representation lays the groundwork for future research on this task.
%A comparison of different approaches to parsing the integrated representation for English shows that joint modeling is beneficial and that the presence of semantic roles actually improves parsing of the rest of the UCCA structure.
\end{abstract}

% linguex examples layout
\setlength{\Exlabelsep}{0.1em}
\setlength{\SubExleftmargin}{1.25em}

%%%%%%%%%%%%%%%%%%%%%%%%%%%%%%%%%%%%%%%%%%%%%%%%%%%%%%%%%%%%%%%%%%%%%%%%%
\section{Introduction}

%%%%%%%%%%%%%%%%%%%%%%%%%%%%%%%%%%%%%%
%%         S L O G A N              %%
%%%%%%%%%%%%%%%%%%%%%%%%%%%%%%%%%%%%%%
%                                    %
% Modular without being incompatible %
%                                    %
%%%%%%%%%%%%%%%%%%%%%%%%%%%%%%%%%%%%%%

%%%%%%%%%%%%%%%%%%%%%%%%%%%%%%%%%%
%%     TODO in Intro            %%
%%%%%%%%%%%%%%%%%%%%%%%%%%%%%%%%%%
%
% * explain how the paper will test the hypothesis that the integration works
%   - quality control of heurstics?
%   - intrinsic and extrinsic ML experiments
%
% * allude to challenges that we encountered during integration
%   - mention contradicting aspects of UCCA and SNACS designs
%
% * compare to other schemes (more detailed in Discussion section)
%
%%%%%%%%%%%%%%%%%%%%%%%%%%%%%%%%%%

A common thread in many approaches to meaning representation 
is the idea that abstract structures can describe semantic invariants 
that hold across paraphrasing or translation: 
for example, semantic dependency relations capturing
predicate-argument structures or other types of semantic relations 
that can be annotated within sentences \citep[e.g.,][]{bohmova2003prague,oepen2015semeval,banarescu2013abstract}.
These annotation schemes can be distinguished by various design principles 
such as language-specificity; 
the level of granularity of meaning elements; 
the reliance on morphosyntactic criteria to define the units of semantic annotation; 
the extent to which human annotators specify semantics from scratch;
and many others \citep{abend-17}.

\begin{figure}[t]
    \centering
    \forestset{
default preamble={
for tree={parent anchor=north, child anchor=north, s sep-=2.1ex, l sep=0, l=2em, font=\sffamily\small}
}
}

\begin{forest}
[
	[, edge label={node[midway,sloped,fill=white,inner sep=1pt,font=\sffamily\tiny]{H}}
	  [I, edge label={node[midway,sloped,fill=white,inner sep=1pt,font=\sffamily\tiny]{A}}, tier=word]
	  [went, edge label={node[midway,sloped,fill=white,inner sep=1pt,font=\sffamily\tiny]{Process}}, tier=word]
	  [, edge label={node(Go)[midway,sloped,fill=white,inner sep=1pt,font=\sffamily\tiny]{A$|$\psst{Goal}}}
		  [to, edge label={node[midway,sloped,fill=white,inner sep=1pt,font=\sffamily\tiny]{R}}, tier=word]
		  [ohm, edge label={node[midway,sloped,fill=white,inner sep=1pt,font=\sffamily\tiny]{Center}}, tier=word]
	  ]
	]
	[after, edge label={node[midway,sloped,fill=white,inner sep=1pt,font=\sffamily\tiny]{Linker}}, tier=word]
	[, edge label={node(Ex)[midway,sloped,fill=white,inner sep=1pt,font=\sffamily\tiny]{H$|$\psst{Explanation}}}
	  [reading, edge label={node[midway,sloped,fill=white,inner sep=1pt,font=\sffamily\tiny]{Process}}, tier=word]
	  [, l sep+=2em, edge label={node[midway,sloped,fill=white,inner sep=1pt,font=\sffamily\tiny]{A}}
	      [some, edge label={node(Qu)[pos=0.56,sloped,fill=white,inner sep=1pt,font=\sffamily\tiny]{Q$|$\psst{Quantity}}}, tier=word]
		  [of, edge label={node[pos=0.6,sloped,fill=white,inner sep=1pt,font=\sffamily\tiny]{R}}, tier=word]
		  [the, edge label={node[pos=0.6,sloped,fill=white,inner sep=1pt,font=\sffamily\tiny]{Fxn}}, tier=word]
		  [reviews, edge label={node[pos=0.6,sloped,fill=white,inner sep=1pt,font=\sffamily\tiny]{Center}}, tier=word]
	  ]
	]
%	[., edge label={node[midway,fill=white,font=\tiny]{U}}, tier=word]
]
% nonterminals
\path[fill=black] (.parent anchor) circle[radius=3pt]
				  (!1.child anchor) circle[radius=3pt]
%				  (!2.child anchor) circle[radius=3pt]
				  (!3.child anchor) circle[radius=3pt]
%				  (!4.child anchor) circle[radius=3pt]
%				  (!11.child anchor) circle[radius=3pt]
%                  (!12.child anchor) circle[radius=3pt]
                  (!13.child anchor) circle[radius=3pt]
%                  (!131.child anchor) circle[radius=3pt]
%                  (!132.child anchor) circle[radius=3pt]
%                  (!31.child anchor) circle[radius=3pt]
                  (!32.child anchor) circle[radius=3pt];
%                  (!321.child anchor) circle[radius=3pt]
%                  (!322.child anchor) circle[radius=3pt]
%                  (!323.child anchor) circle[radius=3pt]
%                  (!324.child anchor) circle[radius=3pt];
%\path[fill=magenta] (!11.child anchor) circle[radius=3pt]
%                    (!21.child anchor) circle[radius=3pt];
%\path[fill=green] (!1.child anchor) circle[radius=3pt];
%\path[fill=cyan] (!2.child anchor) circle[radius=3pt];
%\path[fill=orange] (!13.child anchor) circle[radius=3pt];
%% remote edges
\draw[dotted] (!3.child anchor) to node(I-remote)[midway,sloped,fill=white,inner sep=1pt,font=\sffamily\tiny]{A} (!11.child anchor);
%% lexical supersense edges
\draw[dashed, ->, mdgreen] (!131.child anchor) to (Go);
\draw[dashed, ->, mdgreen] (!2.child anchor) to (Ex);
\draw[dashed, ->, mdgreen] (!322.child anchor) to (Qu);
\end{forest}
    \caption{Semantic parse illustrating the integrated representation proposed here. Solid edges are the UCCA parse's \emph{primary edges}, and the dotted edge is a \emph{remote edge}. Dashed arrows show how SNACS labels (green small caps) have been mapped onto edges of the UCCA structure from the prepositions on which they were originally annotated. 
    % \nss{Nice! Made the tokens a bit bigger; can the horiz space between them be reduced?}\jp{Done.} 
    % \nss{Since most readers won't know UCCA categories, I spelled them out where there was room} 
    The following UCCA categories are abbreviated: A = Participant, R = Relator, H~=~Parallel scene, Q = Quantifier, Fxn = Function.}
    \label{fig:ex-eng}
\end{figure}
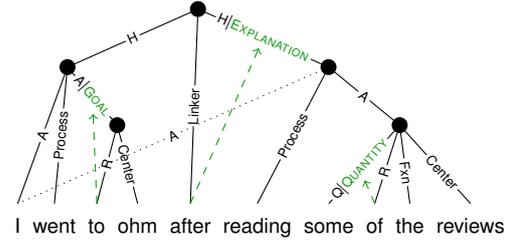

In this work, we seize an opportunity to unite two previously unrelated---yet complementary---meaning representations in NLP. 
On the one hand, Universal Conceptual Cognitive Annotation \citep[UCCA;][]{abend2013universal} 
provides a skeletal structure of semantic units and relations, 
with typologically-based criteria for marking predicate-argument structures, based on ``Basic Linguistic Theory'', an established framework for typological description \citep{Dixon:basic}. 
On the other hand, a recent approach to annotation of English prepositions and possessives \citep[SNACS;][]{schneider2018comprehensive} 
provides an inventory of labels that characterize semantic relations. 
UCCA and SNACS follow similar design principles: 
they are both \emph{language-neutral}, with general-purpose coarse-grained labels rather than lexically-specific senses or roles; 
and they are both designed for \emph{direct} semantic annotation, 
without requiring a syntactic parse as a foundation. 
The philosophy is that these properties will facilitate annotation in many languages and domains that may lack detailed lexicons.
But UCCA makes only the most rudimentary role distinctions, while SNACS 
annotations thus far have not made explicit which elements are being brought into a semantic relation (\cref{sec:bg}).

We propose a design that achieves the best of both worlds, as illustrated for an English example in \cref{fig:ex-eng}. 
Taking advantage of an English corpus that has been separately annotated for both UCCA and SNACS, as well as dependency syntax, we show that the SNACS role labels can be automatically integrated within UCCA structures over a range of syntactic constructions (\cref{sec:integration}).
%(though the work described in this paper is limited to English).
%\jp{removed this because we now have the German case study}
%We use a common corpus to establish the compatibility between the two schemes, which requires adjustments for a number of syntactic phenomena (\cref{sec:integration}).
Then, we use this corpus to test pipelined, multitask, and joint approaches to parsing the integrated representation (\cref{sec:models}).
Our findings (\cref{sec:exp}) set the stage for future English parsers as well as multi- and cross-lingual extensions. 
\Cref{sec:related} situates this work in the broader landscape of computational meaning representations.

Our main contributions are:

\begin{itemize}[leftmargin=10pt]
    \item a typologically-oriented broad-coverage linguistic \textbf{representation} that captures  predicate-argument structure and semantic roles, without reference to any lexicon;
    \item a procedure to \textbf{integrate} UCCA and (token-level) SNACS annotations for particular sentences, mapping the SNACS labels to the appropriate edge in the UCCA structure, by which we create an integrated gold standard; and
    \item initial results for the \textbf{integrated parsing task}, comparing several alternatives that couple the learning\slash prediction of UCCA and SNACS in various ways. We find that optimizing for the two objectives jointly works best.
    % \jp{Omri, does this state the bottom line as you had in mind?}
    %demonstrating how the semantic role information relates to the structural information;  
\end{itemize}
Data and code from these experiments are open-sourced to facilitate future work on this task.%
\footnote{Integrated data: \url{https://github.com/jakpra/ucca-streusle}; parser code: \url{https://github.com/jakpra/tupa}; the integration routine and evaluation scripts are being released as part of the UCCA PyPI package and under \url{https://github.com/jakpra/ucca}.}

\section{Background}\label{sec:bg}

To better understand the distinctions that we expect to be captured by such a framework, consider the following examples:

\ex. 
    \a. Her picture was on the wall.\label{ex:picture}
%    \b. Her house was on fire.\label{ex:house}
%    \c. Her teacher was on the way.\label{ex:teacher}
%    \d. Her wedding was on the mountain.\label{ex:wedding}
    \b. Her speech was on security.\label{ex:speech}

%\oa{Jakob, how about adding ``Her car was on the market'' too? just because ``on the way'' is a MWE which in German happens to be literally translatable. or is this covered by (b)?}
%\jp{Yes, that was my intention. But I'm happy to exchange examples for others that are clearer.}
Despite parallel syntax and overlapping vocabulary, the sentences above vary in numerous aspects of meaning:

\begin{itemize}[leftmargin=10pt]
\item The NPs \textit{her picture} and \textit{the wall}
% in \cref{ex:picture}, \textit{her house} in \cref{ex:house}, and \textit{her teacher} in \cref{ex:teacher} 
denote entities that stand in a certain locative relation to each other, as signaled by the preposition \textit{on}.
%, while \textit{her wedding} and \textit{her speech} denote events directly.
\item 
%The PPs \textit{on the wall} (\cref{ex:picture}) and \textit{on the mountain} (\cref{ex:wedding}) describe the location (non-core circumstantial information) of an entity or event, 
In contrast, the relation between \textit{her speech} (which is an event, not an entity) and \textit{security} is a different one, \textsc{Topic}, despite being signaled by the same preposition.
%, and \textit{(was) on fire} (\cref{ex:house}) and \textit{(was) on the way} (\cref{ex:teacher}) are idiomatic predications, rather than referring to specific scene elements.
%, as well as a 
% and a \textsc{Social} relation in \cref{ex:teacher}.
\end{itemize}

This is made explicit in the German translations of these sentences:

\ex. 
    \ag. Ihr Bild hing \textbf{an} der Wand .\label{ex:picture-de}\\
         \small{Her} \small{picture} \small{hung} \small{\textbf{at}} \small{the} \small{wall} \small{.}\\
%    \bg. Ihr Haus \textbf{brannte} .\label{ex:house-de}\\
%         \small{Her} \small{house} \small{\textbf{burned}} \small{.}\\
%    \cg. Ihr Lehrer war auf dem Weg .\label{ex:teacher-de}\\
%         \small{Her} \small{teacher} \small{was} \small{on} \small{the} \small{way} \small{.}\\
%    \dg. Ihre Hochzeit war auf dem Berg .\label{ex:wedding-de}\\
%         \small{Her} \small{wedding} \small{was} \small{on} \small{the} \small{mountain} \small{.}\\
    \bg. Ihre Rede war \textbf{über} Sicherheit .\label{ex:speech-de}\\
         \small{Her} \small{speech} \small{was} \small{\textbf{over}} \small{security} \small{.}\\

In addition, the possessive pronoun \textit{her} (\textit{ihr}\slash \textit{ihre}) signals a prototypical \textsc{Possession} relation in \cref{ex:picture}\slash \cref{ex:picture-de}, but the core role of \textsc{Agent} in \cref{ex:speech}\slash \cref{ex:speech-de}.
% \oa{I moved this sentence after the De example as both are Ihr in German.}
    
As we can see, the natural lexical choices for expressing the \textsc{Location} relation between the picture and the wall and the \textsc{Topic} relation in German have diverging literal translations to English. 
Thus, the empirical study of cross-linguistic commonalities and differences between form and meaning calls for a common metalanguage to describe the relations between mentioned events and entities, as marked by case and adpositions. %\jp{rephrased a bit to highligh the importance of both schemes}
% \nss{OK? I think the following is less clear:}
% between the speech and security in German.
% It is therefore important to separate the semantic role a linguistic unit expresses in context, which is often preserved across translations, from its form and prototypical lexical function, which is language-specific.

% A lot of research in recent years on creating precise and efficient annotation schemes for full-- or even cross-sentence semantic structure has brought forth several co-existing schemata, of which the Abstract Meaning Representation (AMR) \citep{banarescu2013abstract} and Semantic Dependencies (SD) \citep{oepen2014semeval} are among the more well-known.
% While these two schemes are language-specific and require expert annotators due to high linguistic complexity, an alternative is posed by the Universal Conceptual Cognitive Annotation (UCCA) scheme \citep{abend2013universal}, which emphasizes cross-linguistic applicability and coarse-grained categories that are intuitive to apply, even for non-experts.

% In this work, we leverage the complementary advantages of SNACS\jp{Needs to be introduced (maybe using parts of the commented paragraph below).} and UCCA by integrating them into a shared, multi-layered framework, manifested in a new corpus, which can be used as training data to improve both the SRL and semantic structure parsing subtasks, as well as downstream applications.

Our approach to such a representation consists of utilizing two existing semantic representations. UCCA (\cref{sec:ucca}) captures the structure of predicate-argument and head-modifier relations at a high level, crucially distinguishing units that evoke a \textbf{scene} (event or state) from other units. SNACS (\cref{sec:snacs}) disambiguates semantic roles as signaled by adpositions and possessives, but only directly annotates a function word, without formalizing the semantic relation that it mediates.\footnote{Note that mapping between syntactic and semantic relations varies by construction: in \emph{She spoke on security}, the semantic head of the relation between \emph{spoke} and \emph{security} corresponds to the syntactic head (the verb), whereas in \emph{Her speech was on security}, UD treats \emph{security} as the syntactic head (\cref{sec:integration}).}
%governor and object, or any other sentence-semantic distinctions.\oa{shall we add that when disambiguating SNACS, one needs to obtain approximations of the gov/obj using UD? or maybe too technical for this section..}
Both of these schemas have the guiding principle to be language-independent, eschewing a lexicon and defining a closed inventory of semantic categories.

\subsection{UCCA}\label{sec:ucca}
    
%\nss{this is from a previous paper. rewrite to use example in }
UCCA is a semantic annotation scheme rooted in typological and cognitive linguistic theory. 
It aims to represent the main semantic phenomena in the text, abstracting away
from syntactic forms. UCCA's foundational layer, which is the only layer annotated over text so far,\footnote{\Citet{prange-19} have proposed and piloted a coreference layer that sits above the foundational layer.} reflects a coarse-grained level of semantics that has been shown to be preserved remarkably well across translations \citep{sulem2015conceptual}.
It has also been successfully used for improving text simplification \citep{sulem2018simple}, as well as to the evaluation of a number of text-to-text generation tasks \citep{birch2016hume,sulem2018samsa,choshen2018usim}.

Formally, UCCA structures are directed acyclic graphs over \textbf{units} (nodes covering a subset of tokens).
Atomic units are the leaves of the graph: individual tokens or unanalyzable MWEs.
Nonterminal units represent larger semantic constituents, such as scenes and compositional participants\slash modifiers. 
The example in \cref{fig:ex-eng} has 5~nonterminal units.
Each unit (save for the root) has a single incoming \textbf{primary} edge, and may also have incoming reentrant \textbf{remote} edges to express shared argumenthood. The primary edges of a UCCA structure thus form a tree, which along with the remote edges, forms a DAG.\footnote{There is also the capability to annotate implicit units, but these are ignored in the standard evaluation and we do not address them here.}

Edges are labeled with one or more \textbf{categories} indicating a kind of semantic relationship. 
The small set of categories includes State and Process for static or dynamic scene predicates, respectively; Participant, Time, and Adverbial for dependents of scenes; Center for the head of a non-scene unit (usually an entity); and
Elaborator and Quantity for modifiers of entities. Scenes can be semantic dependents (Participant of another scene, Elaborator of a non-scene). Multiple scenes at the same level are called Parallel Scenes, and connectives between them are Linkers. % not mentioning Connector or Ground, as these are not relevant for examples, or Relator or Function

%UCCA distinguishes scene-evoking from non-scene-evoking nouns (e.g., ``war'' as opposed to ``book''), and dynamic from static scenes. It also identifies many types of multi-word expressions, such as light verbs (e.g., ``take a shower'') and idiomatic predicates (e.g., ``keep to himself'').

UCCA makes a distinction between different functions of prepositions, the most common cases of which are: (1)~phrasal verbs (e.g., ``give {\bf up}''), annotated as internally unanalyzable; (2)~linkers between scenes; e.g., in \cref{fig:ex-eng}, ``after'' links the going to ohm scene, and the reading scene); (3)~main relations in scenes (e.g., ``The apple tree is {\bf in} the garden''); and (4)~case markers within a scene or a participant, or {\it Relators} in UCCA terms (e.g., in 
\cref{fig:ex-eng}, ``to'' and ``of'' are such markers).

However, apart from distinguishing temporal modifiers, the UCCA scheme does not provide any \textbf{semantic role} information: thus the analyses of ``the dark wizard defeated by Gandalf'' and ``the dark wizard's defeat of Gandalf'' are nearly isomorphic---obliterating the distinction between agents and patients in the semantics---though the grammatical encoding of the noun phrases in question (subject, by-PP, possessive, of-PP) leaves no ambiguity about the intended roles to a human reader.

\subsection{SNACS}\label{sec:snacs}

%\nss{from old intro:}We exploit a general-purpose set of role labels inspired by VerbNet's, but reinterpreted and extended for direct comprehensive annotation of English prepositions and possessives \citep[SNACS;][]{schneider2018comprehensive}. 
%Unlike typical approaches to annotating the lexical semantics of predicate-argument structure (including FrameNet, PropBank, and VerbNet), this approach does not require a language-specific predicate lexicon---and is therefore compatible with UCCA's design principle of crosslinguistic applicability.

SNACS is an inventory of 50 roles\slash relations %, which are broadly divided into circumstantial, participant, and configurational senses, and have proven to comprehensively capture relations expressed by 
used to disambiguate adpositions and possessives in multiple languages, including English \citep{schneider2018comprehensive,schneider2017adposition}, Mandarin Chinese \citep{zhu-19}, and to a lesser extent, Korean, Hindi, and Hebrew \citep{hwang-17}.
% nothing published about German
Many of the SNACS labels, such as \psst{Agent}, %\jp{maybe this is just me again, but doesn't the green font give away that this is us?\nss{not necessarily. we could be imitating ourselves :) it matches figure 1 which is useful i think} can we leave it black until the final version? (same for figure 1)}, 
\psst{Theme}, and \psst{Topic}, are derived from VerbNet's \citep{verbnet} core roles of predicates. (Others, such as \psst{Quantity} and \psst{Whole}, are for entity modification.)
But unlike VerbNet, FrameNet \citep{framenet}, and PropBank \citep{propbank}, SNACS does not require a language-specific predicate lexicon (hence \citet{schneider2018comprehensive} use the term ``supersenses'', which we adopt in the remainder of this paper)---and is therefore compatible with UCCA's design principle of crosslinguistic applicability.\footnote{SNACS also annotates the \textbf{function} of a preposition token---its lexical semantics which may be distinct from its semantic role \citep{hwang-17}. Only scene roles are taken into account in the present analysis.}

%\paragraph{Construal analysis.}
% \jp{should we even mention this here? (We don't use functions at all...) If so, could you write a short paragraph, Nathan?} 

% \oa{I wrote a sentence}

%\paragraph{Lexical markables.}
Currently, SNACS labels are applied directly to lexical items,
%In its current state, the SNACS framework focuses on semantic roles expressed by lexical items, 
without marking up underlying structure on either the subword (morphological) or the sentence-structure level.

\section{Automatically Integrating Semantic Roles with UCCA}\label{sec:integration}

% \ex.
%     \a. {[ [ Her$^{S+A}_{\psst{Possessor}}$ (picture)$_A$ ]$_E$ picture ]$_A$ was$_F$ on$_S$ {[ the$_F$ wall$_C$ ]$_{A:\psst{Locus}}$} .}
%     \e. {[ [ Her$_{S+A:\psst{Possessor}}$ (house)$_A$ ]$_E$ house$_S$]$_A$ was$_F$ {[ on fire ]$_{S:\psst{Manner}}$} .}
%     \b. {[ Her$_{A:\psst{SocialRel}}$ teacher$_{P+A}$ ]$_A$ was$_F$ [ on his way ]$_{P:\psst{Manner}}$} .
%     \c. Her$_{A:\psst{Agent}}$ wedding$_P$ was$_F$ {[ on the mountain ]$_{A:\psst{Locus}}$ .}
%     \d. Her$_{A:\psst{Originator}}$ speech$_P$ was$_F$ {[ on security ]$_{A:\psst{Topic}}$ .}

With the benefit of a jointly annotated corpus,
%annotated for UCCA, SNACS, and Universal Dependencies \citep[UD;][]{nivre2016universal} syntax (\cref{sec:data}), 
we examined the data and determined that the proper placement of adpositional semantic role labels is fairly deterministic given certain syntactic patterns and their structural counterparts in UCCA.
Here we present a rule-based method for automatically integrating token-based semantic role annotations from SNACS into an UCCA graph as edge refinements. 
We use these rules to construct a gold standard for analysis and parsing of the integrated representation.
The rules we use, though empirically grounded in the English Web Treebank corpus \citep{ewtb}, make no specific assumptions about language or lexicon,
% \oa{we may want to hedge the term ``universal''; cross-linguistically applicable?}
as they solely depend on UCCA, SNACS, and Universal Dependencies \citep[UD;][]{nivre2016universal} annotation, all of which are frameworks designed to be cross-linguistically applicable.
Thus, we expect the rules could be adapted to other languages with only minor changes if the underlying annotations are applied consistently, though this will require testing in future work.

\subsection{Data}\label{sec:data}

We use the STREUSLE 4.0 corpus \citep{schneider-15,schneider2018comprehensive},
which covers the reviews section from the English\_EWT treebank of
UD 2.3,
and lexical semantic annotations for the same text.\footnote{UD: \url{https://github.com/UniversalDependencies/UD_English-EWT}; however, as described in \cref{sec:exp-setup}, we use automatic dependency parses in all experiments, to emphasize generalizability.}
The same corpus has been annotated with UCCA by \citet{hershcovich2019content}.\finalversion{\footnote{We
do not experiment with other UCCA data despite it being available, as
it is based on an older version of the UCCA guidelines; as the rules and the
EWT annotated corpus are both using UCCA v2, it would be an unfair comparison.\nss{this may no longer be true at camera-ready time, so it would be nice to try wiki data}}}
We use the standard train/dev/test split for this dataset (\cref{tab:quant}).
%\textbf{44804}/5394/5381 tokens in 2406/554/535 sentences.
% \oa{Jakob, are we using just a sub-part? if so, please update the numbers}
% \jp{some of the training isn't annotated yet (at least)}
\Cref{sec:snacs-dist} shows the distribution of linguistic phenomena at issue here.

\subsection{Procedure}\label{sec:heur}

Given an UCCA graph with annotated terminals, the integration routine 
projects a SNACS annotation of a token onto the appropriate edge representing a semantic relation.
This is illustrated by dashed arrows in \cref{fig:ex-eng}. 
The procedure starts with a single terminal node, traversing the graph upwards until it finds an edge that satisfies the criteria given by the rules. 
%Below, we describe these rules and their linguistic motivation.
%
% \nss{Maybe: start with the canonical PPs, and show that they can be syntactically adverbial or adnominal, and the governer can be scene-evoking or not. This could be a 2x2 table of examples. Then have a subsection for all the syntactically weird cases. Quantify in the corpus how many tokens of each.}
%
% \nss{some of the examples have brackets around the full phrase, which I think is distracting}
The rules concern canonical prepositional phrase modifiers, plus a variety of syntactically or otherwise anomalous constructions, such as copulas and adverbs.

% \jp{add pseudo-code in appendix?}\oa{I don't think it's necessary since you release the code}

\subsubsection{Canonical PPs}
% \nss{in general, in this section it is not always obvious which of these statements about structural relationships are in terms of UD and which are in terms of UCCA. e.g.: how do we know if an adposition relates X to Y? is that syntactically defined?}

The adpositional constructions annotated in STREUSLE can be adnominal or adverbial modifiers and arguments, where both the nouns and verbs that are being elaborated on can evoke either scenes or non-scene units in UCCA (\cref{tab:examples}).
First, we take a look at expressions marked with Relators in UCCA, which generally correspond to prototypical syntactic PPs.

% Note that semantic heads and dependents do not always coincide with syntactic governors and objects;
% a frequent trigger of syntax\slash semantics ``inversion'' is \textbf{of}.
% \nss{of-inversion is already a complexity. first discuss the simple case, maybe mentioning ``to ohm'' in \cref{fig:ex-eng}.}\jp{I don't want to give examples here, but just foreshadow what comes in the next two paragraphs. Since of-inversion happens for both scene and non-scene mods, I feature it here, but maybe we can omit this.}\nss{if you want to mention it under scenes and say it also happens in non-scenes, that's fine. but i think it's worth first explaining scene modifiers using the example in fig. 1. the next 2 paragraphs sound very technical}
% % \nss{`governor' and `object' are normally syntactic terms, so we should probably say `semantic head vs. dependent'}\jp{OK}

\begin{table}[t]
    \centering\small
    \begin{tabular}{|c|l|l|}
    
\multicolumn{1}{c}{} & \multicolumn{1}{c}{\textbf{scene}} & \multicolumn{1}{c}{\textbf{non-scene}} \\\hline
\multirow{2}{*}{\textbf{verb}} & I \underline{\textit{went}} [ \textbf{to} ohm ] &  \textit{Quit} [ \textbf{with} the \\
          &    & \underline{overstatements} ] !\\\hline
\multirow{4}{*}{\textbf{noun}} &  Wonderful \underline{\textit{service}}  & Cheapest \textit{drinks}  \\
          & [ \textbf{for} large group ]                         & [ \textbf{in} Keene ]  \\\cdashline{2-3}
          &  [ 10 minutes ] \textbf{of} & [ No amount ] \textbf{of} \textit{sugar} \\
          &  \underline{\textit{paperwork}}   & and milk can \underline{mask} it .  \\\hline
    \end{tabular}
    \caption{Syntactic and semantic dimensions of canonical adpositional phrase constructions. The adposition is \textbf{bolded}, the semantic head is \textit{italicized}, and the semantic dependent is [ bracketed ]. Rows indicate whether the semantic head is nominal or verbal, while columns differentiate between scene-evoking and non-scene-evoking heads. Scene-evokers are \underline{underlined}.}
    % \nss{I'd avoid `ago' because it's syntactically weird and might confuse people. The point is that the PP is predicative. Also, with predicative PPs the preposition is the scene-evoker, right? Maybe underline the scene-evoker?}
    \label{tab:examples}
\end{table}

\paragraph{Modifiers of scenes.} 
%Let us use the term \textbf{scene modifier} for a dependent in a scene, whether a participant or circumstantial modifier such as time or location.
In general, if the adposition marks a modifier of a scene---i.e., the adposition is the first or last terminal in the modifier unit's yield---the supersense should refine the role of that dependent.
%This is easy if the unit \textit{marked}\footnote{I.e., the adposition is the first or last terminal in the unit's yield.} by the adposition is not the scene-evoking unit itself.
The adposition's parent unit is refined by the supersense (``to ohm'' in \cref{fig:ex-eng} and \cref{tab:examples}; ``for large group'' in \cref{tab:examples}).

% \jp{Quantity SS has reversed direction}
% 10 minutes of paperwork
% \ex.
%     \a. a Blue cross has no record of aa reversal . \textit{year} \textbf{of} \underline{negotiations}

% \ex.
%     \a. ``she ate$_P$ [the apple] [\textbf{on$_R$} the patio]$_{A:\psst{Locus}}$''
%     \b. ``she ate$_P$ [the apple] [\textbf{in$_R$} the morning]$_{T:\psst{Time}}$''
%     \c. ``[a year]$_{T:\psst{Quantity}}$ [\textbf{of$_R$} negotiations]$_P$''
%     \d. ``[the cat] emerged$_P$ [\textbf{from$_R$} under$_R$ the couch]$_{A:\psst{Source}}$"\jp{what is this phenomenon called again?}

\paragraph{Modifiers of non-scenes.}
Where the adposition relates a modifying (elaborating or quantifying) unit to a non-scene unit, the supersense refines the modifying unit.
If the adposition is the first or last terminal in a non-Center unit, that unit gets refined (``in Keene'' in \cref{tab:examples}).
This includes the case when the adposition marks the predicate evoking the scene of which the syntactic governor is a modifier (``with the overstatements'': ``Quit'' is treated as an aspectual modifier of the ``overstatements'' event).

For partitive constructions like ``the top \textbf{of} the mountain'', % and related phenomena such as ``late \textbf{at} night'', 
both the syntactic governor and object of the adposition are marked as UCCA Centers, indicating that they are on a \textit{semantically equal} level---neither one is clearly only modifying or being modified by the other. %; the expressions ``modify each other''.
In this case, the supersense refines the syntactic object of the adposition.

\paragraph{Quantities.}
Another special case is where the adposition is labeled as \psst{Quantity}, 
in which case the unit for its syntactic governor%
\footnote{We determine the head noun of the syntactic governor and object using a script released together with the STREUSLE corpus: \url{https://github.com/nert-nlp/streusle/blob/master/govobj.py}. Since semantic UCCA units do not always align with syntactic phrases, we choose the UCCA unit containing the head token of the syntactic governor or object in its yield.} (``10 minutes'' in \cref{tab:examples}) receives the refinement: 
e.g., ``[~some~] \textbf{of} the reviews'' in \cref{fig:ex-eng}; in \cref{tab:examples}, ``[~10 minutes~] \textbf{of} paperwork'' and ``[~no amount~] \textbf{of} sugar'' (the bracketed expression is the \psst{Quantity}).

% then we check whether the construction describes a (temporally) \textit{quantified} scene, as in ``10 minutes \textbf{of} paperwork'', where the supersense of the adposition, \psst{Quantity}, %\nss{why not Duration? can you use the fig. 1 Quantity example instead?}\jp{because that's what's in STREUSLE (in order for it to be Duration, the pobj would have to be the duration). Can't use the Quantity example from figure 1 here because `reviews' is not a scene} 
% actually denotes the role of the syntactic governor

% If it marks a Center unit that has no Center siblings, we check for a \psst{Quantity} supersense to determine whether to refine the syntactic object or governor (``some \textbf{of} the reviews'' in \cref{fig:ex-eng}; also see above).

% \jp{Quantity SS has reversed direction}
% \ex.
    % \a. ``she ate [the apple$_C$ [\textbf{from$_R$} the oldest tree]$_{E:\psst{Source}}$''
    % \b. ``she ate [[a small amount]$_{Q:\psst{Quantity}}$ \textbf{of$_R$} apples$_C$]$_A$''
    % \c. ``she saw [the top$_C$ \textbf{of$_R$} the mountain$_{C:\psst{Whole}}$]$_A$"\label{ex:top}

% \ex.
    % \a. ``[the apple] is \textbf{on$_S$} [the table]$_{A:\psst{Locus}}$''
    % \b. ``she 's \textbf{around}$_{S:\psst{Locus}}$''

\subsubsection{Non-canonical phenomena}

For other, less prototypical constructions involving SNACS-annotated expressions, such as copulas, linked parallel scenes, possessive pronouns, and idiomatic PPs, we have additional rules, summarized in supplementary material (\cref{app:noncanon}).

\begin{table}[t]
    \centering\small
    \begin{tabular}{|l|r|r|r|r|}
\multicolumn{1}{c}{} &  \multicolumn{1}{c}{\textbf{train}}  & \multicolumn{1}{c}{\textbf{dev}}   & \multicolumn{1}{c}{\textbf{test}} & \multicolumn{1}{c}{\textbf{total}} \\\hline
sentences & 2,723   & 554   & 535 & 3,812 \\
tokens    & 44,804  & 5,394 & 5,381 & 55,579 \\\hline
\textbf{SNACS-annotated} & 4,522 & 453 & 480 & 5,455 \\
successful integ. & 4,435 & 447 & 473 & 5,355 \\
matches synt.~obj & 3,924 & 403 & 438 & 4,765 \\\hline
    \end{tabular}
    \caption{Quantitative analysis of adpositional constructions in the corpus.}
    \label{tab:quant}
\end{table}

\begin{table}[t]
    \centering\small
    \begin{tabular}{|l|r|r|r|r|}
\multicolumn{1}{c}{} &  \multicolumn{1}{c}{\textbf{train}}  & \multicolumn{1}{c}{\textbf{dev}}   & \multicolumn{1}{c}{\textbf{test}} & \multicolumn{1}{c}{\textbf{total}} \\\hline
total primary edges & 54,204 & 6,628 & 6,623 & 67,455 \\
total remote edges & 2,881 & 349 & 387 & 3,617 \\\hline
%SNACS-annotated & 4,522 & 453 & 480 & 5,455 \\
\textbf{refined} & 4,473 & 449 & 479 & 5,401 \\\hline
%matches synt.~obj &  & 403 & 438 & 4,765 \\\hline
$\geq 1$ edge refined & 38 & 2 & 6 & 46 \\
remote edges & 33 & 2 & 5 & 40 \\\hline
\textbf{canonical} & 2,468 & 242 & 270 & 2,980 \\
\hspace{1em}scene mod & 2,124 & 219 & 254 & 2,597 \\
\hspace{1em}non-scene mod & 344 & 23 & 16 & 383 \\\hdashline
\textbf{non-canonical} & 2,005 & 207 & 209 & 2,421 \\
%\hspace{1em}scene mod &  &  &  &  \\
%\hspace{1em}non-scene mod &  &  &  &  \\\hdashline
\hspace{1em}predication & 167 & 19 & 23 & 209 \\
\hspace{1em}linkage & 461 & 54 & 41 & 556 \\
\hspace{1em}intransitive adp. & 261 & 19 & 23 & 303 \\
\hspace{1.5em}\scriptsize scn-mod\phantom{0}nscn-mod & \scriptsize 189 \phantom{0} 72 & \scriptsize 14 \phantom{00} 5 & \scriptsize 20 \phantom{00} 3 & \scriptsize 223 \phantom{0} 80 \\
\hspace{1em}approximator & 14 & 0 & 1 & 15 \\
% \hspace{1em}\scriptsize scn-mod\phantom{0}nscn-mod & \scriptsize 0 \phantom{0} 14 & \scriptsize  \phantom{0} & \scriptsize 0 \phantom{00} 1  \\
\hspace{1em}possessive pron. & 897 & 81 & 95 & 1,073 \\
\hspace{1.5em}\scriptsize scn-mod\phantom{0}nscn-mod & \scriptsize 774 \phantom{} 123 & \scriptsize 72 \phantom{00} 9 & \scriptsize 87 \phantom{00} 8 & \scriptsize 933 \phantom{} 140  \\
\hspace{1em}infinitival & 66 & 18 & 11 & 95 \\
\hspace{1.5em}\scriptsize scn-mod\phantom{0}nscn-mod & \scriptsize 15 \phantom{0} 51 & \scriptsize 0 \phantom{0} 18 & \scriptsize 1 \phantom{0} 10 & \scriptsize 16 \phantom{0} 79  \\
\hspace{1em}PP idiom & 139 & 16 & 15 & 170 \\\hline
%\hspace{1em}\scriptsize scn-mod\phantom{0}nscn-mod &  \\\hline
    \end{tabular}
    \caption{Refined UCCA edges by construction type, according to our heuristic. (The non-canonical subcategories are mutually exclusive.)
    %\jp{so are the canonical ones. is this not clear from the indentation? I can add angled arrows or something.}\nss{it's clear from the indentation that these are subsets, and it's clear from the names of the canonical subtypes that they're mutually exclusive. but i wanted to clarify that the canonical subtypes are non-overlapping.}}
    }
    \label{tab:quant-edges}
\end{table}

\subsection{Quantitative Analysis}\label{sec:snacs-dist}

We run the integration routine on our dataset and report statistics in \cref{tab:quant,tab:quant-edges}.
% \jp{should we also provide a confusion matrix over these structural properties and the actual supersenses (subhierarchies would probably be enough), testing the hypothesis that scene mods, predications, and linkages have mostly circumstance and participant roles, non-scene mods have circumstance and configuration roles and possessive pronouns are participants or configurations? Or would that be too linguistically involved for the audience?}\nss{I doubt there will be space to get into a lot of detail on this, but it may be worth mentioning some of these trends in text}
% \jp{agree. I'll put something together tomorrow and then we can see how much we talk about that in the text}
% \nss{test set: 480/6240 primary edges are SNACS-refined. near the top of the table let's count the total numbers of primary edges and remote edges in the UCCA}
% \nss{can we split up table 2? first group of rows counts SNACS tokens; second counts edge refinements. and maybe explain how a SNACS token can contribute to multiple edge refinements}
% \jp{two tables: one in terms of tokens, one in terms of edges}
% \nss{I think the first point is that most adpositions can be integrated}
% The heuristic rules find a relation in the graph that satisfies our criteria %
% \oa{do you mean coverage for the rules?}\jp{yes. should I make that clearer?} 
The heuristic rules have a coverage of 98\% -- 99\% (row `successful integ.'\ divided by row `SNACS-annotated' in \cref{tab:quant}).
88.5\% (train) -- 92.6\% (test) of refined units contain the syntactic complement\footnote{We use the term `syntactic complement' to include head tokens of prepositional objects and subordinate clauses, as determined by the script mentioned above. If no prepositional object is given in the STREUSLE corpus, we consider the adposition itself and only count towards this metric if the refined edge is its incoming preterminal edge.}
% \nss{when I hear `object' I normally assume direct object. Explain this is the preposition's object or complement, and also how this applies to possessives and subordinate clauses}
(\cref{tab:quant}), indicating that while syntax may often give a good approximation to the semantic relations marked by adpositions, a direct mapping from syntactic into semantic structure is not always trivial\longversion{ (see \textbf{Quantities} above)}.

% \nss{example where the object is not contained?}
%it should not be taken
%there is no consistent isomorphism between form and meaning of adpositional constructions.

In \cref{tab:quant-edges}, we see that among the canonically adpositional SNACS targets, the vast majority mark scene modifiers (including  participants).
The various non-canonical targets modify both scenes and non-scenes,\longversion{\footnote{Recall that possessive pronouns, for example, can denote participants in dynamic (\cref{ex:picture,ex:picture-de}) and stative scenes, in addition to their prototypical role as markers of possession (\cref{ex:speech,ex:speech-de}).}} except for linkages and predications, which naturally only operate at the scene-level, and approximators, which only elaborate on non-scenes. 
Similar to canonical adpositional constructions, possessive pronouns and intransitive adpositions have a tendency to modify scenes rather than non-scenes.
% \nss{complicated---remove citation to simplify?:}
Those infinitivals that are not inter-scene Linkers (which are counted under \textit{linkage}) are mostly non-scene modifiers.
% (cf.~``Inherent Purposes'' in the SNACS annotation guidelines \citep{schneider2017adposition}, e.g. ``this is a good place \textbf{to} stay'').
% \jp{maybe I should add scene vs. non-scene numbers for non-canonical}.
% Possessive pronouns, for example, can denote agentive or otherwise active participants (\cref{ex:picture,ex:picture-de}), as well as general or social relatedness (``''), in addition to their prototypical role as markers of possession (\cref{ex:speech,ex:speech-de}).

Remote edges are only rarely affected by the SNACS integration, so we exclude them from evaluation in \cref{sec:exp}.
% \jp{check where they occur}

% \nss{what \% of edges receive a SNACS refinement? i.e. how many primary edges total?}

\subsection{Difficult Cases \& Limitations}\label{sec:diff-cases}

% 98.1 98.7 98.5

Our heuristics are based solely on universal semantic and syntactic annotations, with no assumptions about the grammar or lexicon of any specific language.
However, there are some limitations to the rules that deserve discussion.
%
% \jp{in general, if the UCCA structure is only slightly off, the heuristics will predict some weird edge placement. E.g., they don't check the tag of the edge where a SS ends up, except in a few cases (scene vs. non-scene, center)}
Most importantly, as many rules are highly sensitive to UCCA structure and categories, errors or inconsistencies in human and automatic UCCA annotation are likely to throw the system off.
This can be mitigated with strict constraints and careful reviewing during manual annotation, but cannot be fully avoided when applying the rules to automatically generated UCCA.
% \nss{unnecessary?:}A future extension to the system might calculate a confidence score and refrain from certain predictions if it is not confident enough.

% Furthermore, despite 

% \jp{to formulate:}
% the rules take into account the position (first or last) of an R within its parent unit. If there are languages that have adpositions somewhere in the middle of a unit, that's problematic

% \nss{this paragraph seems in the weeds. maybe it could go in an appendix, but most readers should just know that MWEs are a challenge and a couple of examples where the heuristics fail}
Multiword expressions (MWEs) are another source of difficulty. Both UCCA and STREUSLE mark idiomatic MWEs, but follow slightly different guidelines.
% (e.g., grounding sentence-adverbials such as ``\textbf{to} \textbf{our} dismay'' or nonreferential idioms like ``\textbf{in} the same way'' are marked unanalyzable in UCCA, but are transparent in SNACS; ``in front of'' and ``in town'' are, respectively, a multiword preposition and an idiomatic PP in STREUSLE, but analyzable in UCCA).
The heuristic rules actually recover from most MWE misalignments;
% For SNACS MWEs, only the first token carries the supersense, and  since the graph traversal starts from the preterminal node, SNACS tokens that are part of a larger unanalyzable UCCA unit (i.e., share their preterminal parent with other terminals) are integrated correctly, as long as the UCCA MWE contains no more than one SNACS token.
however, constructions that are MWEs in UCCA and contain multiple SNACS targets, such as \textit{as-as} comparatives, are not fully resolved by our heuristic, as we cannot assign individual edge refinements for the adpositions' competing supersenses, given that we start traversing the UCCA graph from the shared preterminal node.

\section{Models}\label{sec:models}

We hypothesize that our combined (lexical and structural) semantic representation is not only linguistically plausible, but also useful in practice, as the annotations on both levels should be informative predictors for each other.
That is, we expect that knowing what semantic role is signaled by an adposition informs the underlying semantic structure of the sentence, and vice versa.

In order to test this hypothesis, we use our annotated corpus to 
parse into the integrated representation.
% train a novel automatic parser for UCCA that takes the lexical semantic information of the supersenses as features (\cref{sec:ex1}), which we implement by modifying TUPA \citep{hershcovich2017transition}, a transition-based parser for UCCA.
%
% In a second experiment (\cref{sec:ex2}), we train the parser to jointly predict the UCCA structure and supersenses for a given sentence.
%
We consider several different ways of orchestrating
%\oa{nice verb!} 
the prediction of the foundational UCCA structure and the prediction of SNACS roles:
modeling SNACS and UCCA in (i)~a pipeline, (ii)~a multitask setup with shared parameters, (iii)~a single joint model\finalversion{ (see \cref{fig:exp-setup})}.

% \begin{itemize}
%     \item modeling SNACS and UCCA in (i)~a pipeline, (ii)~a multitask setup with shared parameters, (iii)~a single joint model (see \cref{fig:exp-setup}).
%     \item predicting SNACS labels directly on the predicate-argument edge in UCCA (\jp{adjust}\textbf{relation-level}, as shown in edge labels in \cref{fig:ex-eng}), versus predicting SNACS at the word level and then heuristically propagating it up the graph (\textbf{terminal-level}, corresponding to the sources of the arrows in \cref{fig:ex-eng}).
% \end{itemize}

% \jp{moved the paragraph about evaluation to \cref{sec:exp-setup}}
% \jp{also made the system diagram \\finalversion}

\finalversion{
\begin{figure}[t]
    \centering
    \includesvg[width=\columnwidth]{ucca-snacs-exps}
    \caption{Caption\nss{TODO}}
    \label{fig:exp-setup}
\end{figure}
}

% \paragraph{Evaluation.}
% Following \citet{hershcovich2017transition}, we evaluate the predicted UCCA graph as follows:
% A predicted edge is counted as correct if there exists an edge in the reference graph that matches both its label and its terminal yield.
% As there may be different counts of edges in the two graphs, we calculate precision and recall.
% Note that the supersenses are only used as features in this setting; they are not predicted by the parser.

\subsection{Baseline: TUPA}

We choose the neural transition-based graph parser TUPA \citep{hershcovich2017transition,hershcovich2018multitask} as a strong baseline for UCCA parsing. It was the official baseline in the recent SemEval shared task on UCCA parsing \citep{hershcovich2019semeval}.
% \footnote{TUPA has been surpassed in performance by some of the submissions to the shared task. However, as these are still unpublished results that use code that has not yet been released, we use TUPA as our baseline.\nss{it is now released. As of the time we ran our experiments?}\nss{this footnote is forcing an awkward layout. can it be moved?}\jp{maybe we can add a comment to the Related Work section instead, similar to the one about Nelson's SNACS tagger?}}
% \jp{doesn't this identify us as being involved in the shared task? or are the results publicized somewhere?}\oa{there is an arxiv version online \url{http://www.cs.huji.ac.il/~oabend/papers/semeval2019.pdf}}}

TUPA's transition system is defined to address the different formal structural phenomena exhibited by UCCA structures, notably reentrancies and discontiguous units. There are transitions for creating nonterminal nodes, and for attaching terminal and nonterminal nodes to a primary or remote (reentrant) parent with an UCCA category label on the edge. The transition system is general enough to be able to tackle parsing into a variety of formalisms, including SDP \citep{oepen2015semeval} and a simplified form of AMR \citep{banarescu2013abstract}; \citet{hershcovich2018multitask} take advantage of this flexibility in their multitask learning framework.

TUPA's learning architecture largely follows that of \citet{kiperwasser2016simple}. It encodes the parser's configuration (buffer, stack and intermediate graph structure) using BiLSTMs, and predicts the next transition using an MLP, stacked on top of them. 
Token-based features, including POS tags, dependency parses, as well as NER and orthographic features, are embedded in the BiLSTM. 
% from SpaCy.\footnote{\url{http://spacy.io}}
% Pre-trained word embeddings are taken from fastText \citep{bojanowski2017}.\jp{I think we should make sure that this is not too redundant with \cref{sec:exp-setup}}\oa{shall we just omit the spacy and fastText refs here?}\jp{ok}
Another set of features, taking into account the partially constructed graph and previously predicted transition types, is fed into the MLP directly.
% \jp{added remark about token-based/BiLSTM features here}

% \oa{add about the distinction between token and transition features}
% \jp{there are features based on tokens and nodes from both stack and buffer, as well as the last predicted transition, and the token-based ones are embedded in the BiLSTM?}

\subsection{Pipeline}\label{sec:ex1}

We extend TUPA by providing the SNACS label as a feature on the adposition token.\footnote{We report here only results for the setting in which a supersense is added as a feature of the \emph{preposition} token. We also experimented with using it as a feature of the syntactic object token---which often, but not always, heads the semantic object (cf.~\cref{tab:quant})---but got similar or worse results.}
This is added in preprocessing in the same way as the syntactic features listed above (including the BiLSTM encoding).
At testing time, we obtain SNACS labels for automatically identified targets from the SVM model of \citet{schneider2018comprehensive}.

\begin{figure*}[t]
    \centering\small
    \vspace*{-11cm}\hspace*{-1.7cm}
    \includegraphics[scale=0.9]{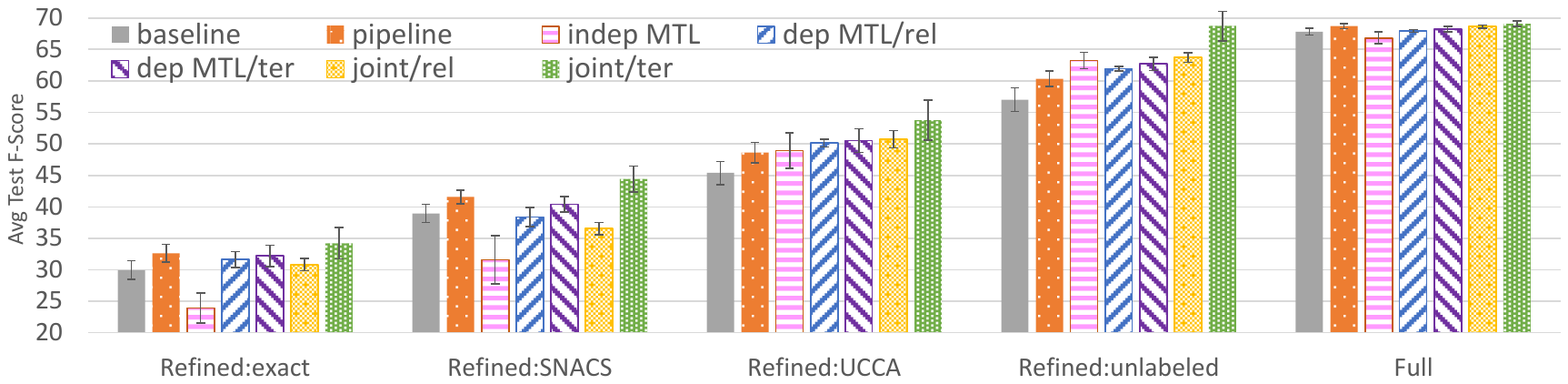}
    \vspace*{-11cm}
    \caption{Average F1-score on the test set over 5 random restarts with error bars indicating standard deviation. \textit{ter} stands for terminal-level and \textit{rel} for relation-level SNACS refinement (prediction or features).}
    \label{fig:results}
\end{figure*}

\begin{table*}[ht]
    \centering\small
    \setlength{\tabcolsep}{4.4pt}
    \begin{tabular}{|c|c|ccc|ccc|ccc|ccc|ccc|}
    % \hline
\multicolumn{2}{c}{\textbf{system}}  & \multicolumn{3}{c}{\textbf{Refined: exact}} & \multicolumn{3}{c}{\textbf{Refined: SNACS}} & \multicolumn{3}{c}{\textbf{Refined: UCCA}} & \multicolumn{3}{c}{\textbf{Refined: unlabeled}} & \multicolumn{3}{c}{\textbf{Full}}   \\
\multicolumn{1}{c}{setup} & \multicolumn{1}{c}{ref}   &    P    &    R    &    \multicolumn{1}{c}{F} &  P    &    R    &    \multicolumn{1}{c}{F} &    P    &    R    &    \multicolumn{1}{c}{F} & P    &    R    &    \multicolumn{1}{c}{F} & P    &    R    &    \multicolumn{1}{c}{F} \\\hline
BL  & \multirow{2}{*}{--} & 30.2  &  29.7  &	30.0  &	39.3  &	38.7  &	38.9  &	45.7  &	45.1  &	45.4  &	57.4  &	56.6  &	57.0  &	68.2  &	67.6  &	67.8  \\
 \scriptsize (oracle SNACS)    &   & \scriptsize 45.4 & \scriptsize	45.1 & \scriptsize	45.2 & \scriptsize	62.7 & \scriptsize	62.3 & \scriptsize	62.5 & \scriptsize	48.9 & \scriptsize	48.6 & \scriptsize	48.7 & \scriptsize	62.7 & \scriptsize	62.3 & \scriptsize	62.5 & \scriptsize	69.5 & \scriptsize	68.9 & \scriptsize	69.2 \\\hline
% \scriptsize (oracle SNACS)    &   & \scriptsize 50.8 & \scriptsize 50.1 & \scriptsize 50.5 & \scriptsize 67.1 & \scriptsize 66.4 & \scriptsize 66.7  & \scriptsize 53.9 & \scriptsize 53.4   & \scriptsize 53.7  & \scriptsize 67.1 & \scriptsize 66.4 & \scriptsize 66.7  &  \scriptsize 71.0  &  \scriptsize 69.4 & \scriptsize 70.2  \\\hline\hline
\multirow{1}{*}{pipeline} & \multirow{2}{*}{ter}  & 32.9  & 32.4  &	32.6  &	42.0  &	41.3  &	41.6  &	49.1  &	48.2  &	48.6  &	60.9  &	59.9  &	60.4  &	68.8  &	68.6  &	68.7 \\
% oracle SNACS
\scriptsize (oracle SNACS) &  & \scriptsize  53.5 & \scriptsize	53.2 & \scriptsize	53.3 & \scriptsize	70.4 & \scriptsize	70.0 & \scriptsize	70.2 & \scriptsize	57.0 & \scriptsize	56.7	 & \scriptsize 56.9 & \scriptsize	70.4 & \scriptsize	70.0 & \scriptsize	70.2 & \scriptsize	71.0 & \scriptsize	70.7 & \scriptsize	70.8 \\\hline

% &   & \scriptsize 52.8 & \scriptsize 52.4 & \scriptsize 52.6 & \scriptsize 70.9 & \scriptsize 70.4 & \scriptsize 70.6 & \scriptsize 56.5 & \scriptsize 55.9 & \scriptsize 56.2 & \scriptsize 70.9 & \scriptsize 70.4 & \scriptsize 70.6 & \scriptsize 70.4 & \scriptsize 69.9 & \scriptsize 70.2  \\\hline
\multirow{1}{*}{indep MTL}                   & ter     &  26.0 &	22.2 &	23.9 &	34.3 &	29.3 &	31.6 &	53.1 &	45.4 &	49.0 &	68.7 &	58.6 &	63.2 &	67.1 &	66.6 &	66.8
  \\\hline
\multirow{2}{*}{dep MTL} & ter     & 34.4     &	30.3     &	32.2     &	43.1     &	38.0     &	40.4     &	53.9     &	47.5     &	50.5     &	66.9     &	59.1     &	62.7     &	68.4     &	68.1     &	68.2
  \\
    & rel     & 32.7     &	30.6     &	31.6     &	39.6     &	37.2     &	38.3     &	51.8     &	48.6     &	50.1     &	64.0     &	60.1     &	61.9     &	68.3     &	67.6     &	67.9
   \\\hline
\multirow{2}{*}{joint} & ter      & \textbf{34.5}     &	\textbf{34.3}     &	\textbf{34.2}     &	\textbf{44.6}     &	\textbf{44.3}     &	\textbf{44.4}     &	53.9     &	\textbf{53.5}     &	\textbf{53.7}     &	69.0     &	\textbf{68.5}     &	\textbf{68.7}     &	\textbf{69.5}     &	\textbf{68.7}     &	\textbf{69.1}
 \\
      & rel     & 34.0     &	28.1     &	30.8     &	40.4     &	33.4     &	36.5     &	\textbf{56.1}     &	46.4     &	50.7     &	\textbf{70.3}     &	58.2     &	63.7     &	69.1     &	68.3     &	68.7
 \\\hline
    \end{tabular}
    \caption{Experimental results, averaged over 5 random restarts. The baseline system (BL) for UCCA is TUPA version 1.3.9 without any modifications, retrained on our data.
    For the sake of generalizability and consistency with our own preprocessing, we use system-predicted SNACS categories from the auto-id/auto-syntax setting from \citep{schneider2018comprehensive} in the BL and pipeline setups.
    Results where the system has access to gold SNACS annotations on adposition tokens are shown in small font.
    }
    \label{tab:results}
\end{table*}

\subsection{Multitask}\label{sec:ex2}
\citet{hershcovich2018multitask} showed that UCCA parsing performance can be improved with multitask learning \citep[MTL;][]{caruana1997multitask} on several semantic and syntactic parsing tasks.
% \nss{this is about using shared representations with other tasks as inductive bias for training the UCCA parser, not actually predicting an integrated structure, right?}
%We make use of this finding and train the parser predict UCCA structure and SNACS supersenses jointly and in a multitask setup.
We examine whether alternately optimizing two objectives, one for UCCA and one for SNACS, leads to mutually favorable biases via shared parameters.
There are multiple ways the two tasks can be orchestrated:
% One approach is to define the SNACS task so that it does not rely on the predicted UCCA structure (we call this \textbf{independent MTL}). 
% Or, we can train the SNACS task in direct interaction with the UCCA parsing task
% % as operating on the relevant edges predicted by the UCCA task
% (\textbf{dependent MTL}).
% \nss{can this be thought of as an extra transition, but trained separately?}\jp{yes, if it helps to conceptualize it that way...but I'm not sure it does.}\oa{I don't have a strong opinion}
% \jp{actually I think we shouldn't call it an extra transition; that would blur the boundaries between MTL and transition-based parsing in a way that people from both areas would probably be confused}

\paragraph*{Independent MTL.}
This is the multitask learning (MTL) setup from \citet{hershcovich2018multitask}, where separate transition classifiers are trained on different tasks simultaneously, sharing and mutually updating the BiLSTM encoding.\footnote{Note that our setup differs from that of \citet{hershcovich2018multitask} in two key points: In contrast to the auxiliary tasks used in the aforementioned work, SNACS prediction as formulated by \citet{schneider2018comprehensive} is not a structured, but a (per-token) classification task (however, as described above, we transform it into an artificially structured task to make it conform with the input format expected by TUPA). Furthermore, we are interested in both UCCA and SNACS performance, expecting both tasks to benefit from each other's complementary semantic content.}
We consider as auxiliary tasks (a)~SNACS scene role classification and (b)~the decision of which UCCA unit is refined by a SNACS-annotated token. % as defined in \cref{sec:integration}.
We encode these tasks as parsing tasks analogous to UCCA parsing as follows: for each training item in (a), we create a graph consisting of a root and up to 4 children: the syntactic governor (if available), the preposition token, the syntactic object (if available)---all of which have dummy edge labels---as well as a dummy terminal carrying the SNACS supersense.
For each training item in (b), we consider the full UCCA structure, but the edge labels are simply boolean values indicating whether an edge is refined or not.

We also train a separate model with SNACS classification as the primary task and UCCA parsing and SNACS integration as auxiliary tasks, whose predictions are integrated in postprocessing for the combined evaluation (\cref{tab:results}), and which is evaluated independently in \cref{tab:snacs-results}.
% \jp{add pointer to \cref{tab:snacs-results}}

% \jp{need two models: one optimized for UCCA, one for SNACS; integration in postprocessing}

\paragraph*{Dependent MTL.}\label{sec:dep-mtl}
Here we train the SNACS task in direct interaction with the UCCA parsing task.
We enhance TUPA with a separate MLP that, given an edge, classifies its supersense refinement (a null category can be chosen to indicate an unrefined edge).
This network is run after each edge-creating transition.
Its input features are the same as for the transition classifier, including the BiLSTM encoding.
Since the two classifiers alternate in making forward passes and updating the shared BiLSTM, they indirectly contribute to each other's input.
% \jp{make this clearer, maybe with a figure}
% \jp{how's this?}
% \paragraph*{}
% Note that our setup differs from that by \citet{hershcovich2018multitask} in two key points:
% In contrast to the auxiliary tasks used in aforementioned work, SNACS prediction as formulated by \citet{schneider2018comprehensive} is not a structured, but a (per-token) classification task.
% Furthermore, we are interested in both UCCA and SNACS performance (instead of treating one as auxiliary to the other), expecting both tasks to benefit from each other's complementary semantic content.
% After training for 100 epochs, we store as the final model the one that maximizes $F_1^{UCCA} + F_1^{SNACS}$. \jp{not sure we still want this}
%We revisit the token-level and the relation-level settings introduced in \cref{sec:ex1}:
Here we have an option of where in the UCCA structure to initially predict the supersense label.
In the \textbf{terminal-level (ter)} setting, we predict supersense refinements only on preterminal edges, and then apply the integration rules (\cref{sec:integration}) as postprocessing.
%of 
%pre-identified \jp{I just realized that I actually didn't include the code I once had for this, so both here and for the joint setup, there is nothing telling the parser at prediction time which tokens should get a supersense. So this is comparable to the auto-id setting in the SNACS paper, but the parser is doing the autoid, not identify.py} 
%adposition tokens, 
In the \textbf{relation-level (rel)} setting, we parse directly into the integrated representation.
%, predicting supersenses as refinements of edges with relevant labels.
To do this, we preprocess the training data with our integration routine.
However, during parsing, there is no explicit restriction that supersense-refined edges must have an adposition token in their yield---thus the model could, in theory, learn to predict adequate role supersenses even when it is not signaled by a lexical marker (though it will get penalized for that in our current evaluation).

\subsection{Joint} %\jp{(composing, concatenating, compound, concatenated?) categories}
Finally, we train a single classifier on the integrated data, concatenating UCCA and SNACS categories, to predict parsing transitions using the new compound categories.
%\finalversion{\nss{add terminal-level results:}We again consider the terminal-refinement and the relation-refinement settings.}
We revisit the terminal-level and the relation-level settings introduced in \cref{sec:dep-mtl}.
% , except that here we have no way of strictly constraining the prediction of supersenses to pre-identified adposition tokens.
% Instead, we use a token-based boolean feature to softly guide the parser towards relevant words.

% \paragraph{Evaluation.} \nss{remove if this is covered by the above}
% UCCA edge tags are evaluated separately from SNACS edge refinements, using the standard UCCA metric described in \cref{sec:ex1}.
% % \jp{In the implementation I got from Adi, a refinement was only counted as correct if both tag and refinement are correct. I changed that so that refinements are evaluated in parallel to tags.}
% % , i.e., an edge is counted as correct if its label is correct and its yield matches the reference yield. 
% We only evaluate refinements that are present (non-null) in either the reference passage or the predicted passage (i.e., we do not count true negatives). 
% \jp{remove until here}

% \jp{this will still hold if we evaluate SNACS on the edge level for these; otherwise we'd just not report SNACS here: }Since we do not, in the evaluation of the joint edge-refinement setup, check which token an edge refinement originally came from, the results of this experiment are not directly comparable with the lexical supersense classifiers by \citet{schneider, gonen}.
% % \jp{E.g., linkers project their SS onto their sibling that contains the obj, so that sibling would not contain the supersensed token in its yield.}

\section{Experiments}\label{sec:exp}

\subsection{Experimental Setup}\label{sec:exp-setup}

\paragraph{Preprocessing.}
We follow \citet{hershcovich2018multitask} in obtaining automatic POS and NE tags, as well as syntactic dependency relations using SpaCy 2.0, and pretrained word vectors from fastText.\footnote{\url{https://spacy.io/}; \url{https://fasttext.cc/}}
For all setups that use or predict SNACS supersenses, we include the gold standard scene role categories for pre-identified targets from the STREUSLE 4.0 corpus in our training and development data.
In the test data we identify adposition targets using the heuristics introduced in \citet{schneider2018comprehensive}.
For the joint prediction setup (\cref{sec:ex2}), we also include the head terminal of the syntactic governor and object for each adposition as features, using the same heuristics as in \cref{sec:integration}.
% \jp{gold-ID for adpositions on train and dev, auto-id on test; (syntactic) govobj. maybe mention that the UCCA integration heuristics are like a semantic govobj (currently only for obj, but gov should be possible; in fact, in a prior version of the script I also output the governing UCCA unit)}

\paragraph{Architecture and hyperparameters.}

For classifying the next transition, TUPA uses a multi-layer perceptron (MLP) with 2 hidden layers and a softmax output layer, on top of a 2-layer {BiLSTM} \citep{kiperwasser2016simple}.
Building on previous work, we train for 100 epochs, using the stochastic gradient descent optimizer for the first 50, and AMS-grad \citep{reddi2018ams} for the remaining 50 epochs,\footnote{Except for the independent MTL setting, where we stop training after the first 50 epochs.} and apply early stopping post-hoc by keeping the model with the highest performance on the dev set as the final model.
% \footnote{Note that, in the dependent MTL settings, while the SNACS classifier has its own loss function, the early stopping criterion is only based on the labeled UCCA score.}

\begin{table*}[t]
    \centering\small
    \setlength{\tabcolsep}{4.4pt}
    \begin{tabular}{|c|c|rrr|rrr|}
    % \hline
\multicolumn{2}{c}{\textbf{system}}  & \multicolumn{3}{c}{\textbf{UCCA labeled}}  & \multicolumn{3}{c}{\textbf{UCCA unlabeled}}  \\
\multicolumn{1}{c}{setup}                          & \multicolumn{1}{c}{ref}  &    \multicolumn{1}{c}{P}    &    \multicolumn{1}{c}{R}    &    \multicolumn{1}{c}{F} &    \multicolumn{1}{c}{P}    &    \multicolumn{1}{c}{R}    &    \multicolumn{1}{c}{F} \\\hline
BL  & -- & 72.5 \scriptsize \textcolor{gray}{$\pm 0.6$} &	71.9 \scriptsize \textcolor{gray}{$\pm .4$}   &	72.2 \scriptsize \textcolor{gray}{$\pm 0.4$}   &	88.5 \scriptsize \textcolor{gray}{$\pm 0.3$}   &	87.6 \scriptsize \textcolor{gray}{$\pm .5$}   &	88.0 \scriptsize \textcolor{gray}{$\pm .3$}  \\\hline\hline
\multirow{1}{*}{pipeline}                   & \multirow{2}{*}{ter}  &  72.4 \scriptsize \textcolor{gray}{$\pm 0.3$}   &	\textbf{72.0} \scriptsize \textcolor{gray}{$\pm .6$}   &	72.2 \scriptsize \textcolor{gray}{$\pm 0.4$}   &	88.3 \scriptsize \textcolor{gray}{$\pm 0.2$}   &	\textbf{87.8} \scriptsize \textcolor{gray}{$\pm .6$}   &	88.1 \scriptsize \textcolor{gray}{$\pm .4$} \\
% oracle SNACS
\scriptsize (oracle SNACS) &  &  \scriptsize 73.0 \textcolor{gray}{$\pm 0.4$} &  \scriptsize	72.6 \textcolor{gray}{$\pm .7$} &  \scriptsize	72.8 \textcolor{gray}{$\pm 0.5$} &  \scriptsize	88.8 \textcolor{gray}{$\pm 0.2$} &  \scriptsize	88.3 \textcolor{gray}{$\pm .7$} &  \scriptsize	88.5 \textcolor{gray}{$\pm .4$} \\\hline
% % &  &  \scriptsize 72.5 &  \scriptsize 71.8 &  \scriptsize 72.2 &  \scriptsize 88.4 &  \scriptsize 87.5 &  \scriptsize 87.9  \\\hline
\multirow{1}{*}{indep MTL}                   & ter     &  71.0 \scriptsize \textcolor{gray}{$\pm 1.2$} &	70.2 \scriptsize \textcolor{gray}{$\pm .9$} &	70.6 \scriptsize \textcolor{gray}{$\pm 1.0$} &	87.9 \scriptsize \textcolor{gray}{$\pm 1.0$} &	86.9 \scriptsize \textcolor{gray}{$\pm .8$} &	87.3 \scriptsize \textcolor{gray}{$\pm .7$}  \\\hline
\multirow{2}{*}{dep MTL} & ter     &	71.8 \scriptsize \textcolor{gray}{$\pm 0.4$}   &	71.3 \scriptsize \textcolor{gray}{$\pm .2$}   &	71.5 \scriptsize \textcolor{gray}{$\pm 0.3$}   &	88.1 \scriptsize \textcolor{gray}{$\pm 0.4$}   &	87.6 \scriptsize \textcolor{gray}{$\pm .2$}   &	87.8 \scriptsize \textcolor{gray}{$\pm .2$}  \\
                         & rel     & 71.8 \scriptsize \textcolor{gray}{$\pm 0.3$}   &	71.0 \scriptsize \textcolor{gray}{$\pm .1$}   &	71.4 \scriptsize \textcolor{gray}{$\pm 0.1$}   &	88.0 \scriptsize \textcolor{gray}{$\pm 0.3$}   &	87.0 \scriptsize \textcolor{gray}{$\pm .4$}   &	87.5 \scriptsize \textcolor{gray}{$\pm .3$}  \\\hline
\multirow{2}{*}{joint} & ter      & \textbf{72.8} \scriptsize \textcolor{gray}{$\pm 0.3$}   &	71.9 \scriptsize \textcolor{gray}{$\pm .4$}   &	\textbf{72.3} \scriptsize \textcolor{gray}{$\pm 0.3$}   &	\textbf{88.7} \scriptsize \textcolor{gray}{$\pm 0.3$}   &	87.6 \scriptsize \textcolor{gray}{$\pm .4$}   &	\textbf{88.2} \scriptsize \textcolor{gray}{$\pm .3$}  \\
                      & rel     & 72.5 \scriptsize \textcolor{gray}{$\pm 0.4$}   &	71.5 \scriptsize \textcolor{gray}{$\pm .2$}   &	72.0 \scriptsize \textcolor{gray}{$\pm 0.3$}   &	88.5 \scriptsize \textcolor{gray}{$\pm 0.2$}   &	87.2 \scriptsize \textcolor{gray}{$\pm .3$}   &	87.8 \scriptsize \textcolor{gray}{$\pm .2$} \\\hline
    \end{tabular}
    \begin{tabular}{|ccc|}
    % \hline
 \multicolumn{3}{c}{\textbf{SNACS}} \\
   \multicolumn{1}{c}{P}    &    \multicolumn{1}{c}{R}    &    \multicolumn{1}{c}{F}  \\\hline
%  & - & Wiki     &  & 49.1 &	47.5 &	48.3 &	43.2 &	12.9 &	19.9  &          &          &        \\
% BL  & - &          &          &    \\
% \multicolumn{2}{|c|}{\citet{schneider2018comprehensive}}          &    &     &    &     &     &    &     &     &     &          &          &   \\
 58.5 & 58.3 & 58.4  \\\hline\hline
 $-$ & $-$ & $-$ \\
 $-$ & $-$ & $-$ \\\hline
 48.2 \scriptsize  \textcolor{gray}{$\pm 5.6$} &	41.4 \scriptsize  \textcolor{gray}{$\pm 4.8$} &	44.6 \scriptsize \textcolor{gray}{$\pm 5.2$} \\\hline
 60.1 \scriptsize \textcolor{gray}{$\pm 1.8$}   &	53.3 \scriptsize \textcolor{gray}{$\pm 2.0$}   &	56.5 \scriptsize \textcolor{gray}{$\pm 1.9$}  \\
  $-$ & $-$ & $-$ \\\hline
 \textbf{60.5} \scriptsize \textcolor{gray}{$\pm 3.0$}   &	\textbf{60.3} \scriptsize \textcolor{gray}{$\pm 3.4$}   &	\textbf{60.4} \scriptsize \textcolor{gray}{$\pm 3.2$}  \\
  $-$ & $-$ & $-$ \\\hline
    \end{tabular}
    \caption{Results on the respective tasks of UCCA parsing and token-level SNACS prediction, averaged over 5 random restarts, with standard deviation reported next to each average. The baseline system (BL) for UCCA is TUPA version 1.3.9 without any modifications, retrained on our data. The SNACS baseline system is the SVM classifier of \citet{schneider2018comprehensive}.
    % \nss{stdevs add visual noise. make them gray?}\nss{vertical layout problem with `pipeline (oracle SNACS)'}
    %\jp{add stddev?}
    % \textit{ter} stands for terminal-level and \textit{rel} for relation-level SNACS prediction\slash features.
    }
    \label{tab:ucca-results}
    \label{tab:snacs-results}
    \label{tab:ucca-snacs-results}
\end{table*}

\paragraph{Evaluation.}
% For our main evaluation in \cref{tab:results}, we compare our systems along five metrics: a \textbf{full structure score} which evaluates precision and recall of units whose (primary edge) label is the UCCA category label concatenated with the SNACS label, if applicable; 
% an \textbf{UCCA score} which ignores the SNACS labels; an \textbf{unlabeled UCCA score} which disregards all unit labels, considering only the structure; 

For our main evaluation in \cref{sec:results}, we compare our systems along five new metrics: a \textbf{full structure} score which evaluates precision and recall of all units and requires both the UCCA and SNACS categories to be correct, where applicable; 
and \textbf{refined UCCA}, \textbf{SNACS}, \textbf{exact}, and \textbf{unlabeled} scores which only consider SNACS-refined units.
Here, the integrated representation obtained via the rule-based integration (\cref{sec:heur}) serves as the ground truth.
% a gold \jp{silver?} standard.
% \jp{say that we evaluate against the automatically integrated gold (silver?) standard, so the recall for the Refined metrics is always out of 479 (\cref{tab:quant-edges})}
% ; and a \textbf{refined unlabeled} score which merely indicates the precision and recall of graph edges with a SNACS refinement.%
% \jp{as I said in the email, I think token-level makes more sense as it is comparable to other SNACS classifiers. If you think it's valuable, we could switch to edge-level for the integrated output systems.}\nss{Either way, as long as it's a fair comparison. For the integrated output systems, is there a postprocessing step that projects the label onto a token?}\jp{I don't think it can be done deterministically in all cases}.
We also report the standard labeled and unlabeled UCCA scores.\footnote{All of the above metrics are F-scores over the edges, as in \citet{hershcovich2017transition}. 
% The full structure score requires both the UCCA and SNACS categories to be correct.
% \jp{This sentence seems unnecessary. Either we explain birefly how it is computed or not. But I don't like calling it a `trivial extension' and then not saying how it is done.} of the labeled UCCA score.
% \nss{Do we score remote edges?}\jp{we can if we want}\jp{do we?}\nss{how many adpositional units have remote edges---put in table 2? if it's not many then no need to report remote scores}\oa{agree}
}
In addition, for systems which predict a terminal-level SNACS label (before it is mapped to a higher relation in postprocessing), 
we compare SNACS disambiguation performance against \citep{schneider2018comprehensive} in \cref{sec:snacs-results}.
% (both \cref{tab:ucca-snacs-results}).

\subsection{Integrated parsing results}\label{sec:results}

% \jp{from email: So the gold\_ss story is: a) All the systems, especially indep-MTL, are slightly better than baseline at parsing UCCA units that contain SNACS tokens in the gold analysis. Both precision and recall improve. Selected F1 scores: BL=93.1, Indep-MTL=96.0, dep-MTL/ter=95.2. b) The pipeline and especially dep-MTL/ter improve labeled UCCA performance (both precision and recall) for such units. F1 BL=74.1, pipeline=76.4, dep-MTL/ter=77.5. c) There is little effect on overall labeled or unlabeled UCCA (the differences are no more than .6 F1 points above or below baseline).}

% \jp{point to the right tables}
% \jp{adjust to new results}

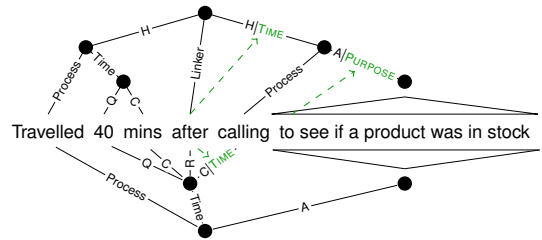
\begin{figure}[t]
    \centering\small
    \forestset{
default preamble={
for tree={parent anchor=north, child anchor=north, s sep-=2.1ex, l sep=0, l=1.6em, font=\sffamily\small}
}
}
\scalebox{0.9}{
\begin{forest}
[
    [,edge label={node[midway,sloped,fill=white,inner sep=1pt,font=\sffamily\tiny]{H}}
        [Travelled, edge label={node[midway,sloped,fill=white,inner sep=1pt,font=\sffamily\tiny]{Process}}, tier=word]
        [, l sep+=0.8em, edge label={node[midway,sloped,fill=white,inner sep=1pt,font=\sffamily\tiny]{Time}}
            [40, edge label={node[pos=0.6,sloped,fill=white,inner sep=1pt,font=\sffamily\tiny]{Q}}, tier=word]
            [mins, edge label={node[pos=0.6,sloped,fill=white,inner sep=1pt,font=\sffamily\tiny]{C}}, tier=word]
        ]
    ]
    [after, edge label={node[midway,sloped,fill=white,inner sep=1pt,font=\sffamily\tiny]{Linker}}, tier=word]
    [,edge label={node(time)[midway,sloped,fill=white,inner sep=1pt,font=\sffamily\tiny]{H$|$\psst{Time}}}
        [calling, edge label={node[midway,sloped,fill=white,inner sep=1pt,font=\sffamily\tiny]{Process}}, tier=word]
        [, edge label={node(purp)[midway,sloped,fill=white,inner sep=1pt,font=\sffamily\tiny]{A$|$\psst{Purpose}}}
            [to see if a product was in stock, roof, tier=word]
        ]
    ]
]
\path let \p1 = (.parent anchor) in node(scene-m) at (\x1, -3.1) {};
% \path let \p1 = (!1.child anchor) in node(1-m) at (\x1,-4.2) {};
\path let \p1 = (!2.child anchor) in node(T-m) at (\x1,-2.4) {};
% \path let \p1 = (!31.child anchor) in node(31-m) at (\x1,-4.2) {};
\path let \p1 = (!32.child anchor) in node(A-m) at (\x1,-2.4) {};
\path let \p1 = (!11.child anchor) in node(travelled-m) at (\x1,-1.8) {};
\path let \p1 = (!121.child anchor) in node(40-m) at (\x1,-1.8) {};
\path let \p1 = (!122.child anchor) in node(mins-m) at (\x1,-1.8) {};
\path let \p1 = (!2.child anchor) in node(after-m) at (\x1,-1.8) {};
\path let \p1 = (!31.child anchor) in node(calling-m) at (\x1,-1.8) {};
% \path let \p1 = (!31.child anchor) in node(tostock-m) at (\x1,-2.6) {};
% \path let \p1 = (!11.child anchor) in node(travelled-m) at (\x1,-2.8) {};
% \draw (root-m) to node[midway,sloped,fill=white,inner sep=1pt,font=\sffamily\tiny]{H} (scene-m);
\draw (scene-m) to node[midway,sloped,fill=white,inner sep=1pt,font=\sffamily\tiny]{Process} (travelled-m);
\draw (scene-m) to node[midway,sloped,fill=white,inner sep=1pt,font=\sffamily\tiny]{Time} (T-m);
\draw (scene-m) to node[midway,sloped,fill=white,inner sep=1pt,font=\sffamily\tiny]{A} (A-m);
\draw (T-m) to node[midway,sloped,fill=white,inner sep=1pt,font=\sffamily\tiny]{Q} (40-m);
\draw (T-m) to node[midway,sloped,fill=white,inner sep=1pt,font=\sffamily\tiny]{C} (mins-m);
\draw (T-m) to node[midway,sloped,fill=white,inner sep=1pt,font=\sffamily\tiny]{R} (after-m);
\draw (T-m) to node(time-m)[midway,sloped,fill=white,inner sep=1pt,font=\sffamily\tiny]{C$|$\psst{Time}} (calling-m);
\draw let \p1 = (calling-m), \p2 = (A-m.north) in (\x2, {\y2 + 3}) to ({\x1 + 12}, {\y1 - 3});
\draw let \p1 = (calling-m), \p2 = (A-m.north) in (\x2, {\y2 + 3}) to ({\x1 + 123}, {\y1 - 3});
\draw let \p1 = (calling-m) in ({\x1 + 12}, {\y1 - 3}) to ({\x1 + 123}, {\y1 - 3});
% nonterminals
\path[fill=black] (.parent anchor) circle[radius=3pt]
                  (!1.child anchor) circle[radius=3pt]
				%   (!2.child anchor) circle[radius=3pt]
				  (!3.child anchor) circle[radius=3pt]
%				  (!4.child anchor) circle[radius=3pt]
%				  (!11.child anchor) circle[radius=3pt]
%                  (!12.child anchor) circle[radius=3pt]
                  (!12.child anchor) circle[radius=3pt]
%                  (!131.child anchor) circle[radius=3pt]
%                  (!132.child anchor) circle[radius=3pt]
%                  (!31.child anchor) circle[radius=3pt]
                  (!32.child anchor) circle[radius=3pt]
                %   (!321.child anchor) circle[radius=3pt]
%                  (!322.child anchor) circle[radius=3pt]
%                  (!323.child anchor) circle[radius=3pt]
%                  (!324.child anchor) circle[radius=3pt]
                %   (root-m) circle[radius=3pt]
                  (scene-m) circle[radius=3pt]
                %   (1-m) circle[radius=3pt]
                  (T-m) circle[radius=3pt]
                  (A-m) circle[radius=3pt];
%\path[fill=magenta] (!11.child anchor) circle[radius=3pt]
%                    (!21.child anchor) circle[radius=3pt];
%\path[fill=green] (!1.child anchor) circle[radius=3pt];
%\path[fill=cyan] (!2.child anchor) circle[radius=3pt];
%\path[fill=orange] (!13.child anchor) circle[radius=3pt];
%% remote edges
% \draw[dotted] (!3.child anchor) to node(I-remote)[midway,sloped,fill=white,inner sep=1pt,font=\sffamily\tiny]{A} (!11.child anchor);
% %% lexical supersense edges
\draw[dashed, ->, mdgreen] (!2.child anchor) to (time);
\draw[dashed, ->, mdgreen] let \p1 = (!31.child anchor) in ({\x1 + 20}, {\y1}) to (purp);
\draw[dashed, ->, mdgreen] (after-m) to (time-m);
\end{forest}
}
    \caption{Simplified sample output. The \textit{joint\slash ter} system (top) generates the intended scene structure and \psst{Purpose} modifier attachment. The \textit{BL} system (bottom) does not, and misses the \psst{Purpose} role altogether.}
    \label{fig:results-ex}
\end{figure}

Our MTL and joint systems outperform the baseline and a feature pipeline on refined UCCA units (\cref{fig:results,tab:results}).
The main benefit from considering UCCA and SNACS together in training is that the parser is better at recovering the (unlabeled) structure of units that should receive a SNACS relation in the integrated representation.
This is illustrated in \cref{fig:results-ex}.
This trend is confirmed in the precision and recall of UCCA units that have a gold SNACS token in their yield (unlabeled F-score: \textit{BL} = 93.1, \textit{dep-MTL/ter} = 95.2, \textit{indep-MTL} = 96.0, see \cref{tab:gold-ss-eval} in the supplementary material).
To the extent that these units are syntactic constituents (see \cref{tab:quant}),
this suggests that multitask learning with syntactic auxiliary tasks \citep{swayamdipta-18,hershcovich2018multitask} might be particularly beneficial for SNACS-augmented UCCA parsing.\finalversion{\nss{I think it is worth looking at the kinds of units that the BL gets wrong but \textit{dep MTL} gets right. E.g. maybe canonical PPs are easy but \textit{dep MTL} is better at getting noncanonical units.}}
% \jp{Change to fit new numbers: While \textit{dep-MTL\slash rel} and \textit{joint\slash ter}, respectively, perform best on UCCA and SNACS categories of refined units, \textit{dep-MTL\slash ter} gets the most exact UCCA+SNACS combinations right.}
The feature pipeline is competitive, but noisy features from a previous classification step limit its performance on refined units.
The upper bounds given by the oracle setting indicate that SNACS features are generally beneficial.
\textit{Indep-MTL} and systems that parse directly into the relation-refined representation struggle with predicting the correct SNACS refinements---and thus also exact UCCA+SNACS combinations---while the \textit{joint/ter} model is consistently the most accurate.

% \begin{table}
%     \centering\small
%     \begin{tabular}{c}
%         [H [P Travelled] [T [Q 40] [C mins] [R after] [C$|$Time calling] ] [A to see if a product was in stock ] ] ] [U .] \\
%         [H [P Travelled] [T [Q 40] [C mins] ] ] [L after] [H$|$Time [P calling] [A$|$Purpose to see if a product was in stock ] ] ] [U .] \\
%          [H [A IMPLICIT] [P Travelled] [T [Q 40] [C mins] ] ] [L after] [H$|$Time [A IMPLICIT] [P calling] [A$|$Purpose to see if a product was in stock ] ] ] [U .] \\
%     \end{tabular}
%     \caption{Caption}
%     \label{tab:results-ex}
% \end{table}

% \begin{figure}[t]
%     \centering\small
%     \vspace*{-1cm}
%     \includegraphics[scale=0.3]{uccasnacs-multiple-runs-learning-detail.pdf}
%     \vspace*{-1.2cm}
%     \caption{}
%     \label{fig:my_label2}
% \end{figure}

% Terminal-level dependent MTL outperforms all other systems on the refined units, while the \textit{joint} prediction system has the highest overall score for our new representation, which is intuitive since it is the only system that directly optimizes for this metric (though it is the worst SNACS predictor).
% Also, predicting SNACS supersenses at the token-level and integrating them in postprocessing (indep\slash terminal-level dependent MTL) is more reliable than parsing into the joint representation directly.

% The indepentend MTL system performs best in the standard UCCA metrics and, together with the oracle-SNACS pipeline setup, is the only system that beats the labeled UCCA baseline, if only by a small margin.
% Most of our systems (except non-orcale pipeline and relation-level dependent MTL) beat the unlabeled UCCA baseline 
However, there is little effect on overall labeled and unlabeled UCCA scores (\cref{tab:ucca-results}).
% That UCCA performance goes slightly down in the pipeline scenario shows that noisy features from a previous classification step can hurt the learning process.
% \jp{Is this still true: The upper bounds given by the oracle setting indicate that SNACS features are generally beneficial.}
Predicting SNACS simultaneously or interactively with UCCA (\textit{joint/rel} and \textit{dep-MTL}) apparently makes the parsing task harder.
Note that particularly in the \textit{dep-MTL} setting, erroneous decisions in one task could negatively affect the other.

\subsection{Token-based SNACS prediction results}\label{sec:snacs-results}

% \jp{add a sentence about joint}
Since some of our systems predict SNACS labels at the terminal level, they are directly comparable to previous work on SNACS classification.
We compare against the auto-id\slash auto-syntax baseline from \citet{schneider2018comprehensive} in \cref{tab:snacs-results}.\footnote{Due to the diverging guidelines on multiword units in UCCA (\cref{sec:diff-cases}), we ignore MWE boundaries.}
Both the \textit{dep MTL} and \textit{joint} systems outperform the baseline in precision; and the \textit{joint} system also in recall, leading to the overall best performance.
% We find that the \textit{joint} setup slightly outperforms the baseline in precision and F-score. %
% (Note that while the SNACS classifier has its own loss function, the early stopping criterion is only based on the labeled UCCA score.)
The \textit{indep-MTL} system does not reach baseline performance.

% \begin{table}[t]
%     \centering\small
%     \begin{tabular}{|c|ccc|}
%     % \hline
% % \multicolumn{1}{c}{}  & \multicolumn{3}{c}{\textbf{SNACS}} \\
% \multicolumn{1}{c}{\textbf{system}} &   P    &    R    &    \multicolumn{1}{c}{F}  \\\hline
% %  & - & Wiki     &  & 49.1 &	47.5 &	48.3 &	43.2 &	12.9 &	19.9  &          &          &        \\
% % BL  & - &          &          &    \\
% % \multicolumn{2}{|c|}{\citet{schneider2018comprehensive}}          &    &     &    &     &     &    &     &     &     &          &          &   \\
% b & 58.5 & 58.3 & 58.4  \\\hline\hline
% \multirow{1}{*}{i}                    & 48.2\scriptsize $\pm 5.6$ &	41.4\scriptsize $\pm 4.8$ &	44.6\scriptsize $\pm 5.2$
%  \\\hline
% \multirow{1}{*}{d}                    & 60.1\scriptsize $\pm 1.8$   &	53.3\scriptsize $\pm 2.0$   &	56.5\scriptsize $\pm 1.9$
%   \\\hline
% \multirow{1}{*}{j} & \textbf{60.5}\scriptsize $\pm 3.0$   &	\textbf{60.3}\scriptsize $\pm 3.4$   &	\textbf{60.4}\scriptsize $\pm 3.2$
%   \\\hline
%     \end{tabular}
%     \caption{Token-based SNACS evaluation, averaged over 5 random restarts. %\jp{add stddev?}
%     The baseline system is the SVM classifier of \citet{schneider2018comprehensive}.
%     }
%     \label{tab:snacs-results}
% \end{table}

\begin{table}[t]
    \centering\small
    \begin{tabular}{|c|c|c|c|}
% \multicolumn{2}{c}{\textbf{system}}  & \multicolumn{1}{c}{} \\
\multicolumn{1}{c}{\textbf{setup}} & \multicolumn{1}{c}{\textbf{ref}}  &  \multicolumn{2}{c}{\textbf{\# params}} \\\hline
BL & $-$ & \multicolumn{2}{c|}{78.9M} \\\hline\hline
pipeline & ter & \multicolumn{2}{c|}{78.9M} \\\hline
\multirow{2}{*}{indep MTL} & \multirow{2}{*}{ter} & \scriptsize{\textbf{UCCA}} & \scriptsize{\textbf{SNACS}} \\
 &  & 82.8M & 81.8M \\\hline
\multirow{2}{*}{dep MTL} & ter & \multicolumn{2}{c|}{79.4M} \\
 & rel & \multicolumn{2}{c|}{79.4M} \\\hline
\multirow{2}{*}{joint} & ter & \multicolumn{2}{c|}{79.2M} \\
 & rel & \multicolumn{2}{c|}{79.2M} \\\hline
    \end{tabular}
    \caption{Number of parameters of each model.}
    \label{tab:params}
\end{table}

% \begin{figure}[t]
%     \centering\small
%     \vspace*{-3.8cm}\hspace*{-1cm}
%     \begin{tikzpicture}
%     \node (img1) {\includegraphics[scale=0.43]{uccasnacs-multiple-runs-learning.pdf}};
%     \draw (-2.5,2.25) -- (0.23,2.25);
%     \draw (-2.5,2.25) -- (-2.5,0.45);
%     \draw (0.23,2.25) -- (0.23,0.45);
%     \draw (0.23,0.45) -- (-2.5,0.45);
%     %
%     \draw (-2.5,0.45) -- (-2.24,-1.85);
%     \draw (0.23,2.25) -- (1.36,0.72);
%     \node[xshift=-5.2cm,yshift=5.5cm] (img2) at (img1.south east) {\includegraphics[scale=0.2]{uccasnacs-multiple-runs-learning-detail.pdf}};
%     \end{tikzpicture}
%     \vspace*{-4cm}
%     \caption{Averaged learning rates on dev set during the first 50 epochs of models optimizing for UCCA. Between epochs 7 and 21, \textit{indep MTL} is able to learn faster than other systems, whereas \textit{dep MTL\slash ter} has a relatively flat learning curve.}
%     \label{fig:learning}
% \end{figure}

\subsection{Model capacity}
We examine whether the differences in performance can really be attributed to the linguistic information in our data or merely to more powerful models by inspecting the number of each model's parameters (\cref{tab:params}).
While we observe some variance in model capacity, we consider these to be minor differences. An exception is the \textit{independent MTL} setup, which consists of two independent models, each dedicated to a specific task. However, this does not seem to give it an advantage in terms of final performance.
The baseline has the fewest parameters, and the overall best condition, \textit{joint/ter}, is neither the smallest nor the largest model, suggesting that the particular linguistic signals and the method of using them have a genuine effect on performance.

\section{Related Work}\label{sec:related}

The benefits of integrating lexical analysis and sentence semantic or syntactic structure have been pursued by a vast body of work over the years. 
% Due to space limitations, we will only touch on a few. %Integration takes many guises. 
Compositional approaches to the syntax-semantics interface, such as CCG \citep{steedman2000syntactic} and HPSG \citep{pollard1994head}, usually integrate the lexicon at the leaves of the syntactic parse, but propagate grammatically-relevant features up the tree.   
A different approach is taken by OntoNotes \citep{hovy2006ontonotes}, which consists of a number of separate, albeit linked tiers, including syntactic and argument structure, but also the lexical tiers of word senses and coreference. 

Role semantics frequently features in structured semantic schemes. Some approaches, such as PropBank and AMR \citep{propbank,banarescu2013abstract}, follow a lexical approach.  The Prague Dependency Treebank tectogrammatical layer \citep{bohmova2003prague} uses a few lexicon-free roles, but their semantics is determined by virtue of their linkage to a lexicalized valency lexicon.
Universal Decompositional Semantics \citep{white-16} instead defines roles as a bundle of lexicon-free features, elicited by crowdsourcing.

% \finalversion{\nss{if room:}
% \jp{Do you want to rearrange things somewhat to make room for this?Nah. It's pretty tangential. We don't focus on MWEs.Yeah ok.}
% Another line of work seeks to integrate multiword expressions into syntactic treebanks, either linking the syntactic structure to an MWE  \citep[e.g.,][]{bejcek-10}, or annotating MWEs as an inseparable part of the scheme (as in AMR). 
% See \citep{rosen2015survey} for a survey of MWE annotation in treebanks. 
% Integration of the two phenomena facilitates jointly parsing them, which has been shown to potentially benefit both MWE identification and parsing \citep{constant-17}.
% }

%\paragraph{Roles and syntax/semantics}
%\begin{itemize}
%    \item see grant proposal
%\end{itemize}

% \oa{re UCCA parsing: I don't think it fits here; we can say TUPA is the only published parser so we experiment with it but it should be in the experimental setup section}

The specific inventory for preposition\slash possessive relations that we use is SNACS, but there is a wider history of disambiguation of these items, especially in English: disambiguation systems have been described for possessives \citep{moldovan-04,badulescu-09,tratz-13}, prepositions with lexicalized sense definitions \shortlong{\citep[e.g.,][]{litkowski-07,tratz-11}}{\citep{litkowski-07,ye-07,tratz-09,hovy-10,tratz-11,clematide-13,gong-emnlp-18}}, and prepositions with coarse-grained classes \citep{ohara-03,ohara-09,srikumar-13,gonen2016semi}. Such disambiguation has also been investigated in tandem with semantic role labeling and parsing \citep{dahlmeier-09,srikumar-11,gong-naacl-18}.
Preliminary work suggests that SNACS may be applicable to subjects and objects, not just PPs, and thus in the future this framework could be extended to all UCCA participants \citep{shalev-19}.

State-of-the-art results on UCCA parsing and SNACS disambiguation are described in contemporaneous work by \citet{jiang-19,liu2019linguistic}, who achieve substantial gains using the ELMo and BERT contextualized word embeddings \citep{elmo,bert}.
This is an orthogonal direction to the one we pursue here, and combining the two is left to future work.

\section{Conclusion}

We have introduced a new representation combining UCCA semantic structures and SNACS adpositional semantic roles; automatically merged existing annotations to create a gold standard; and experimented with several alternatives for parsing the integrated representation.
Our results show that models profit from having access to both structural and lexical semantic information, confirming our hypothesis that UCCA and SNACS are complementary and compatible.

Based on preliminary results from a German corpus, we conjecture that this approach is applicable to other languages with no or only minimal changes---a direction we will explore further in future work.
In addition, we plan to investigate the utility of the enhanced representation for downstream tasks involving meaning-preserving linguistic variation.
% \jp{maybe mention proof-of-concept experiment with German?}\oa{I think you could mention that in a sentence.}

\section*{Acknowledgments}
We would like to thank Daniel Hershcovich and Adi Shalev for letting us use their code and helping fix bugs; Sean MacAvaney and Vivek Srikumar for assistance with computing resources; as well as three anonymous reviewers for their helpful comments and suggestions.
This research was supported in part by NSF award IIS-1812778 and grant 2016375 from the United States--Israel Binational Science Foundation (BSF), Jerusalem, Israel.
% \jp{Nathan, can you put the appropriate grant(s) in here?}
% \jp{I guess it's the BSF UCCA-SNACS grant, and then whatever is funding my travel. Maybe also the one that got us the GPUs?}

% The acknowledgments should go immediately before the references.  Do not number the acknowledgments section ({\em i.e.}, use \verb|\section*| instead of \verb|\section|). Do not include this section when submitting your paper for review.

%\bibliography{acl2018}
\bibliography{uccasnacs}
\bibliographystyle{acl_natbib}

\clearpage

% \maketitle

\appendix

\begin{table*}[bp]
    \centering\small
    \setlength{\tabcolsep}{4.4pt}
    \begin{tabular}{|c|c|ccc|ccc||ccc|ccc|}
    % \hline
\multicolumn{2}{c}{\textbf{system}}  & \multicolumn{3}{c}{\textbf{gold-ss labeled}}  & \multicolumn{3}{c}{\textbf{gold-ss unlabeled}} & \multicolumn{3}{c}{\textbf{gold-ss sib labeled}}  & \multicolumn{3}{c}{\textbf{gold-ss sib unl.}}  \\
\multicolumn{1}{c}{setup}                          & \multicolumn{1}{c}{ref}  &      P    &    R    &    \multicolumn{1}{c}{F} &    P    &    R    &    \multicolumn{1}{c}{F} &    P    &    R    &    \multicolumn{1}{c}{F} &    P    &    R    &    \multicolumn{1}{c}{F} \\\hline
BL  & -- & 74.2 & 73.9 & 74.1 & 93.3 & 92.9 & 93.1 & 47.7 & 48.0 & 47.9 &60.7 & 61.1 & 60.9  \\\hline\hline
\multirow{2}{*}{pipeline}                   & \multirow{2}{*}{ter}  &  76.5  & 76.2 & 76.4 & 94.5 & 94.2 & 94.4 & 48.1 & 49.2 & 48.8 & 61.1 & 62.4 & 61.7  \\
% oracle SNACS
% &  & \scriptsize 73.0 & \scriptsize	72.6 & \scriptsize	72.8 & \scriptsize	88.8 & \scriptsize	88.3 & \scriptsize	88.5 \\\hline

&  & \scriptsize 79.2 & \scriptsize 78.9 & \scriptsize 79.1 & \scriptsize 95.8 & \scriptsize 95.4 & \scriptsize 95.6 & \scriptsize 51.0 &  \scriptsize 50.8 &  \scriptsize 50.9 &  \scriptsize 63.5 & \scriptsize 63.3 &  \scriptsize 63.4  \\\hline
\multirow{1}{*}{indep MTL}                   & ter     & 72.4 & 72.4 & 72.4 & \textbf{96.0} & \textbf{96.0} & \textbf{96.0} & 46.8 & \textbf{50.4} & 48.6 & 59.8 & \textbf{64.5} & 62.1   \\\hline
\multirow{2}{*}{dep MTL} & ter     &	\textbf{77.5} &	\textbf{77.5} &	\textbf{77.5} & 95.2 & 95.2  &  95.2 & 46.7 & 48.9 & 47.8 & 60.2 & 63.2 & 61.6  \\
                         & rel     & 73.5  & 72.9  & 73.2  & 94.1 & 93.3 & 93.7 &\textbf{48.7} & 50.0 & \textbf{49.3} & \textbf{62.4} & 64.1 & \textbf{63.2}  \\\hline
\multirow{2}{*}{joint} & ter      & 73.2 & 73.1 & 73.1 & 95.2 & 95.0 & 95.1 & 47.9 & 46.9 & 47.4 & 62.2 & 60.8 & 61.5 \\
                      & rel     & 73.1  & 73.1 & 73.1  & 95.2  & 95.2 & 95.2 & 44.5 & 46.3 & 45.4 & 57.7 & 59.9 & 58.8 \\\hline
    \end{tabular}
    \caption{UCCA parsing performance on units with a gold SNACS label in their yield (\textit{gold-ss}) and siblings of such units (\textit{gold-ss sib}).}
    \label{tab:gold-ss-eval}
\end{table*}

\section{Non-canonical phenomena requiring special rules}\label{app:noncanon}

\paragraph{Predication.}
If the adposition denotes a predication via a copular construction as in ``They were \textbf{back} quickly'' or ``was not \textbf{in} the agreement as well'', the supersense refines the syntactic object, if it is explicit, or the adposition unit itself, if not.

\paragraph{Linkage.}
Subordinating conjunctions, which are included as adpositional expressions %in STREUSLE 
if the same lemma can also be used as a canonical adposition (with a nominal object), project their supersense onto the subordinated clause (``\textbf{after} [reading some of the news]'' in \cref{fig:ex-eng}).

\paragraph{Infinitival purpose markers.}
While infinitival \textit{to} is not considered as marking a semantic role in most cases, certain uses of \textit{to}, which elaborate on the purpose of an action or entity, are annotated. 
% (Cf. ``Inherent Purposes'' in the SNACS annotation guidelines \citep{schneider2017adposition}, e.g. ``this is a good place \textbf{to} stay''.)
These project their supersense onto the elaborating unit in which they occur. % (``to stay'').

\paragraph{Intransitive adpositions.}
If the adposition is intransitive---due to blurry definitions, we avoid the notions of \textit{particles} and \textit{adverbs} here---the supersense continues to refine the adposition unit (``drive 10 minutes more \textbf{down} to Stevens Creek'').

% \ex. ``she jumped$_P$ \textbf{up}$_{D:\psst{Direction}}$''

\paragraph{Approximators.}
Approximating adpositions like \textit{around} and \textit{about} (``I bought \textbf{about} half of the furniture'') similarly keep their supersense; in contrast to \textit{intransitive} adpositions, which express a relation between their governor and an implicit object, approximators elaborate on their object without specifying a relation towards the governor.

\paragraph{Possessive pronouns.}
Possessive pronouns keep their supersense refinement, whether they express true possession (``we got \textbf{our} food'') or a participant or other derived relation (``\textbf{our} company was just getting started'').
% \nss{unnecessary detail?:}Note that, in contrast to intransitive adpositions, which usually relate to an \textit{implicit}, i.e., unrealized but more or less salient argument, possessive pronouns \textit{explicitly} encode both the role marker (genitive case) and the marked object (pronominal referent) at the same time.

% \ex.
%     \a. ``\textbf{our}$_{A:\psst{Agent}}$ wedding$_P$''
%     \b. ``he likes [\textbf{his$_{A:\psst{SocialRel}}$} teacher$_{P+A}$]''

\paragraph{Idiomatic PPs.}
When a prepositional phrase is classified as idiomatic in STREUSLE (e.g., ``in town''), its supersense refines the adposition token.
% \nss{example}

% \ex. ``good place [\textbf{to$_F$} stay (place)]$_{E:\psst{Characteristic}}$''

% \section{Detailed Hyperparameters}

% \begin{table}[ht]
%     \centering
%     \centering\small
%     \setlength{\tabcolsep}{4.4pt}
%     \begin{tabular}{|l|r|r|}
%         max iterations &  \\
%          & 
%     \end{tabular}
% %     \begin{tabular}{|c|c|ccc|ccc|}
% %     % \hline
% % \multicolumn{2}{c}{\textbf{system}}  &   \\
% % \multicolumn{1}{c}{setup}                          & \multicolumn{1}{c}{ref}  &      P    &    R    &    \multicolumn{1}{c}{F} &    P    &    R    &    \multicolumn{1}{c}{F} \\\hline
% % BL  & -- & \textbf{73.0} & 71.3 & 72.1 & 88.7 & 86.7 & 87.6  \\\hline\hline
% % \multirow{2}{*}{pipeline}                   & \multirow{2}{*}{ter}  &  72.0  & 71.3 & 71.7 & 87.9 & 87.0 & 87.4  \\
% % % oracle SNACS
% % &  & \scriptsize 72.5 & \scriptsize 71.8 & \scriptsize 72.2 & \scriptsize 88.4 & \scriptsize 87.5 & \scriptsize 87.9  \\\hline
% % \multirow{1}{*}{indep MTL}                   & ter     & 72.9 & 71.5 & \textbf{72.2} & \textbf{89.3} & 87.6 & \textbf{88.4}    \\\hline
% % \multirow{2}{*}{dep MTL} & ter     &	71.8 &	71.4 &	71.6 & 88.1 & \textbf{87.7}  &  87.9   \\
% %                          & rel     & 72.5  & 71.4  & 71.9  & 87.8 & 86.6 & 87.2   \\\hline
% % \multirow{2}{*}{joint} & ter      & 72.5 & \textbf{71.6} & 72.1 & 88.5 & 87.5 & 88.0  \\
% %                       & rel     & 72.1  & 71.3 & 71.7  & 88.3  & 87.3 & 87.8 \\\hline
% %     \end{tabular}
%     \caption{Caption}
%     \label{tab:hyperparams}
% \end{table}

\section{Extended Evaluation}

\subsection{Gold SNACS units}
See \cref{tab:gold-ss-eval} for a detailed UCCA parsing evaluation of units with a gold SNACS supersense in their yield and siblings thereof.
Looking at the siblings of SNACS-bearing units can tell us that getting PP unit spans right also helps with non-PP children of a scene.
Furthermore, adpositions that are inter-scene Linkers in UCCA are siblings to their refined arguments.
This might also be a reason why the numbers are generally lower for siblings, as getting a full scene unit correct is more difficult than smaller units. Another reason could be that the set of units that are siblings to gold-ss units is more prone to change due to small differences in the gold-ss units themselves.

\end{document}